\documentclass[sn-mathphys,Numbered,iicol]{sn-jnl}

\usepackage{graphicx}%
\usepackage{multirow}%
\usepackage{amsmath,amssymb,amsfonts}%
\usepackage{amsthm}%
\usepackage{mathrsfs}%
\usepackage[title]{appendix}%
\usepackage{xcolor}%
\usepackage{textcomp}%
\usepackage{manyfoot}%
\usepackage{booktabs}%
\usepackage{algorithm}%
\usepackage{algorithmicx}%
\usepackage{algpseudocode}%
\usepackage{listings}%

\usepackage[edges]{forest}
\usepackage{tabularx}
\usepackage{multirow}
\usepackage{makecell}
\usepackage{colortbl}
\usepackage{tabularray}
\UseTblrLibrary{booktabs}

\newcommand{\representationheading}[1]{\par\smallskip\noindent\textbf{#1}\quad}

\definecolor{BackgroundRed}{RGB}{245,227,217}
\definecolor{BackgroundYellow}{RGB}{255,253,233}
\definecolor{BackgroundBlue}{RGB}{242,242,255}
\definecolor{ColorA}{RGB}{255,0,0}
\definecolor{ColorB}{RGB}{0,0,255}

\forestset{
	title/.style={
		rotate=90,
		anchor=center,
		font=\small
	},
	subtitle/.style={
		minimum width=4.5em,
		align=center,
		fill opacity=.6,
		font=\tiny
	},
	subsubtitle/.style={
		minimum width=5em,
		align=center,
		fill opacity=.6,
		font=\tiny
	},
	methods/.style={
		text width=30em,
		align=left,
		fill opacity=.6,
		font=\fontsize{4pt}{6pt}\selectfont
	},
	figuretitle/.style={
		rotate=90,
		anchor=center,
		align=center,
	},
	figurebranch/.style={
		align=center,
		minimum width=7em
	},
	figureleaf/.style={
		minimum width=18em,
		fill=white
	},
	figuremethod/.style={
		align=left,
		font=\tiny,
		fill=white,
		inner xsep=3pt,
		inner ysep=2pt,
		text width=12em
	}
}

\theoremstyle{thmstyleone}%
\theoremstyle{thmstyletwo}%

\theoremstyle{thmstylethree}%

\begin{document}
	
	\title[Representation Learning in Diffusion and Flow-based Model]{Representation Learning in Diffusion and Flow-based Model: An Application Aspect}
	
	
	\author[1]{\fnm{Yanchen} \sur{Xu}}\email{yanchenxu.tj@gmail.com}
	
	\author[2]{\fnm{Sida} \sur{Huang}}\email{sidahuang2001@gmail.com}
	
	\author[1]{\fnm{Zhenyu} \sur{Gu}}\email{25110890001@m.fudan.edu.cn}
	
	\author[2]{\fnm{Ruishu} \sur{Zhu}}\email{zhuruishu0848@gmail.com}
	
	\author[2]{\fnm{Yilan} \sur{Gao}}\email{2022yl@mail.nwpu.edu.cn}
	
	\author*[3]{\fnm{Hongyuan} \sur{Zhang}}\email{hyzhang98@gmail.com}
	
	\affil[1]{\orgdiv{Fudan University, Shanghai 200433, China}}
	
	\affil[2]{\orgdiv{School of Artificial Intelligence, OPtics and ElectroNics (iOPEN)}, \orgname{Northwestern Polytechnical University, Xi'an 710072, Shaanxi, China}}
	
	\affil[3]{\orgdiv{The University of Hong Kong, Hong Kong, China}}
	
	
	\abstract{Diffusion models and flow-based models have recently become the dominant paradigms in generative modeling, largely due to their ability to learn rich, multi-level visual representations through large-scale training. This creates a bidirectional relationship between generative models and representation learning: improving representation learning enhances generation quality, while the learned representations can be leveraged for broader understanding tasks. This survey systematically explores this interplay with a focus on applications. We propose a three-tier progressive framework that organizes existing works from three perspectives: using representation learning to improve generative capabilities, exploiting generative models to extract representations for perception tasks, and ultimately moving toward general-purpose unified applications. We systematically categorize representative methods across a wide range of downstream tasks, including image classification, dense visual prediction, instance-level perception, and annotation-scarce scenarios. By providing a unified taxonomy and identifying key challenges, this survey aims to clarify the underlying logic of current research and suggest promising directions for future exploration. We hope this work can serve as a valuable reference for researchers interested in harnessing the representation power of generative models for applications beyond generation.}

	
	
	
	\maketitle
	
	\section{Introduction}
	Diffusion models have recently become the dominant paradigm in generative modeling, achieving remarkable success across a wide range of domains, including image synthesis, video generation, audio generation, and molecular design. Meanwhile, flow-based models, as an emerging class of generative models, learn deterministic transports along probability paths, offering unique advantages in inference efficiency and flexibility, and are gradually becoming a compelling complement or even alternative to diffusion models. The success of both model families stems from their ability to learn rich visual representations at multiple levels of abstraction through large-scale training. This naturally raises a question: \textit{\textbf{can these representations be effectively leveraged?}} Furthermore, \textit{\textbf{can such leverage extend beyond generation tasks to broader understanding tasks?}}
	
	Around this question, we find that diffusion models and flow-based models share a tight bidirectional relationship with representation learning and visual understanding. On one hand, the training process of generative models inherently involves learning structured representations of data; improving their representation learning capability can directly lead to more realistic sample generation. On the other hand, these learned representations can also serve downstream understanding tasks such as classification, segmentation, and detection. Notably, diffusion models and flow-based models do not explicitly extract semantic representations. Unlike models such as variational autoencoders, their internal representations are distributed across different network layers, different timesteps, and even different attention heads, and the extraction methods vary depending on the task. In recent years, a surge of research has emerged around representation learning with diffusion models and flow-based models, exploiting the interplay between the two through various paradigms.

	However, these works lack a unified taxonomy, making it difficult for researchers to keep pace with current developments.
	To the best of our knowledge, only one existing survey is exclusively dedicated to this topic, and it focuses solely on diffusion models\cite{survey}. Other related reviews on generative models touch upon representation learning only in passing,  rather than treating it as a central theme\cite{survey2}.
	To effectively organize this growing body of work and clarify its underlying logic, we provide a comprehensive overview and taxonomy of relevant approaches.
	In this survey, we cover representative works published from early 2020s to early 2026, with a selection focus on those that introduce novel representation learning mechanisms, demonstrate clear application value, or define new paradigms in diffusion and flow-based generative models.
	Specifically, this paper introduces existing work from three progressive perspectives: leveraging representation learning to improve generative capabilities, exploiting diffusion and flow-based models for representation learning, and ultimately moving toward generalization and unification. The main contributions of this paper are as follows:
	\begin{itemize}
		\item \textbf{A comprehensive survey perspective.} Unlike existing surveys that focus solely on diffusion models, this paper discusses diffusion models and flow-based models in parallel, systematically reviewing the bidirectional interplay between these two mainstream generative model families and representation learning.
		
		\item \textbf{A progressive taxonomy.} This paper proposes a three-tier progressive framework: from leveraging representation learning to enhance generative capabilities, to exploiting generative models to extract representations for perception tasks, and ultimately toward general-purpose applications and unified frameworks. This framework reveals the intrinsic logic and evolutionary trends of research in this field.
		
		\item \textbf{Discussion of challenges and future directions.} This paper summarizes the key challenges faced by related fields, identifies the significant potential of diffusion models and flow-based models in representation learning, and suggests possible directions for further development.
	\end{itemize}
	\section{Background}
	This section provides the background knowledge for the subsequent chapters. We begin with the basic mathematical frameworks of diffusion models, flow-matching models, and rectified flows (Sec. \ref{Preliminaries}). Then, we introduce the commonly used backbone network architectures for these models (Sec. \ref{Network_Architecture}).
	\subsection{Preliminaries}\label{Preliminaries}
	Let the training data $X$ be sampled from an unknown data distribution $p(x)$. The goal of generative models is to learn the distribution and generate new samples from it. Diffusion models map Gaussian noise to $p(x)$ through progressive denoising, while flow-based methods learn deterministic transport paths from a source distribution to $p(x)$ to accomplish generation.
	\subsubsection{Diffusion Models}
	Diffusion models are a class of latent variable models inspired by considerations from nonequilibrium thermodynamics\cite{DDPM}. The core idea is to learn a forward diffusion process that gradually adds Gaussian noise to data samples, transforming them into Gaussian distribution. The trained neural network generates new samples by reversing the denoising process.
	
	Given a training sample $x_0\sim p(x)$, the forward diffusion process can be defined as a Markov chain:
	\begin{equation}
		p(x_t|x_{t-1}) = \mathcal{N}(x_t;\sqrt{1-\beta_t}x_{t-1},\beta_tI), t = 1,\cdots,T,
	\end{equation}
	where $x_t$ denotes the sample obtained after adding noise at step $t$, $T$ denotes the number of diffusion time steps, and $\beta_t$ is a pre-defined variance schedule. Due to the fact that the conditional distributions at each step are mutually independent, it holds that
	\begin{equation}
		p(x_t|x_0) = \mathcal{N}(x_t;\sqrt{\bar{\alpha}_t}x_0, (1-\bar{\alpha}_t)I),
	\end{equation}
	where $\bar{\alpha}_t = \prod^t_{i=1}(1-\beta_i)$. Hence, with a reparameterization trick, a noisy sample at an arbitrary timestep $t\in\{1,2,\cdots,T\}$ can be sampled directly as:
	\begin{equation}
		x_t = \sqrt{\bar{\alpha}_t}x_0 + \sqrt{1-\bar{\alpha}_t}\varepsilon_t,\quad\varepsilon_t\sim\mathcal{N}(0,I).
	\end{equation}
	In practice, a network $\varepsilon_\theta$ is trained to predict the noise $\varepsilon_t$ given diffusion timestep $t$ and the noisy sample $x_t$. For simplification of implementation, \cite{DDPM} proposes training $\varepsilon_\theta$ with the loss function as follow:
	\begin{equation}
		\mathcal{L} = \mathbb{E}_{t,x_0,\varepsilon_t}||\varepsilon_t - \varepsilon_\theta(x_t,t)||^2.
	\end{equation}
	The above derivation is based on the DDPM framework with discrete timesteps. When the number of diffusion steps $T$ tends to infinity, the discrete diffusion process can be generalized to a continuous-time Stochastic Differential Equation (SDE) formulation\cite{Diffusion-SDE}:
	\begin{equation}
		dx = f(x,t)dt + g(t)dw,
	\end{equation}
	where $w$ is the standard Wiener process, while $f(\cdot, t)$ and g(t) are drift coefficient and diffusion coefficient function, respectively. There are two common choices of SDE formulation. The first one is Variance-Preserving (VP) SDE. It is defined by $f(x,t)=-\frac{1}{2}\beta(t)x, g(t)=\sqrt{\beta(t)}$, where $\beta(t)$ is the continuous generalization of the discrete variance schedule $\beta_t$. Note that VP-SDE corresponds to the continuous limit of DDPM, with the variance remains bounded throughout the diffusion process.The second one is Variance-Exploding (VE) SDE, which allows the variance to grow continuously as $t$ increases. Corresponding to the continuous limit of NCSN\cite{NCSN}, VE-SDE is defined by $f(x,t)=0,g(t)=\sqrt{\frac{d}{dt}\sigma^2(t)}$.
	
	Regardless of which SDE formulation is adopted, Anderson\cite{Reverse-time-SDE} proves that the diffusion process has a reversible time reversal, with the reverse-time SDE as:
	\begin{equation}
		dx = [f(x,t)-g(t)^2\nabla_x\log p(x;t)]dt+g(t)dw,
	\end{equation}
	where $\nabla_x\log p(x;t)$ is the score function which is approximated by a neural network.
	
	Furthermore, diffusion models can operate in either pixel space or latent space\cite{StableDiffusion}, the latter compresses data into a low-dimensional latent space via a pre-trained VAE before performing diffusion, significantly reducing the computational cost of high-resolution image synthesis and providing a more efficient interface for downstream perception task adaptation.
	\subsubsection{Flow Matching Models}
	Given a source distribution $p_0(x)$, typically a standard Gaussian distribution, Flow Matching Models aims to learn a deterministic transport process that smoothly maps $p_0(x)$ to $p(x)$\cite{FM}.
	
	In the flow matching framework, the transport process is modeled by a time-dependent vector field $v(x,t)$, which corresponds to an Ordinary Differential Equation (ODE) as:
	\begin{equation}
		dx_t = v(x_t,t)dt,\qquad t\in[0,1].
	\end{equation}
	The ODE defines a flow $\phi_t$ that pushes the source distribution $p_0$ forward to the target distribution $p_1\triangleq p$, i.e., $p_t=[\phi_t]*p_0$. Given the target probability path $p_t(x)$ and the corresponding vector field $u_t(x)$, a neural network $v_\theta(x,t)$ is trained to regress the vector field:
	\begin{equation}
		\mathcal{L}_{FM} = \mathbb{E}_{t,x}||v_\theta(x,t)-u_t(x)||^2.
	\end{equation}
	However, in practice, we do not have access to $p_t(x)$ and $u_t(x)$. To address this, Lipman et al. \cite{FM} introduced the Conditional Flow Matching (CFM) construction. CFM defines the conditional probability path $p_t(x|x_1)$ of the data sample $x_1$, such that it equals the source probability $p_0$ at time $t=0$, and be a distribution concentrated around $x_1$ at time $t=1$:
	\begin{equation}
		p_t(x)=\int p_t(x|x_1)q(x_1)dx_1.
	\end{equation}
	And the marginal vector field can be formulated as
	\begin{equation}
		u_t(x) = \int u_t(x|x_1)\frac{p_t(x|x_1)q(x_1)}{p_t(x)}dx_1.
	\end{equation}
	$\mathcal{L}_{FM}$ shares the same gradients with respect to $\theta$ with the objective of CFM as follow:
	\begin{equation}
		\mathcal{L}_{CFM}=\mathbb{E}_{t,q(x_1),p_t(x|x1)}||v_t(x)-u_t(x|x_1)||^2.
	\end{equation}
	Considering conditional probability paths of the form \begin{equation}
		p_t(x|x_1)=\mathcal{N}(x|\mu_t(x_1),\sigma_t(x_1)^2I),
	\end{equation}
	the vector field has the form:
	\begin{equation}
		u_t(x|x_1)=\frac{\sigma_t'(x_1)}{\sigma_t(x_1)}(x-\mu_t(x_1))+\mu_t'(x_1).
	\end{equation}
	Two representative designs are as follows:
	\begin{itemize}
		\item \textbf{Diffusion Path}: $\mu_t(x_1)=\alpha_(1-t)x_1$ and $\sigma_t(x_1)=\sqrt{1-\alpha_{1-t}^2}$;
		\item \textbf{Optimal Transport Path}: $\mu_t(x_1)=tx_1$ and $\sigma_t(x_1)=1-(1-\sigma_{min})t$.
	\end{itemize}
	\subsubsection{Rectified Flow}
	Being an important variant of flow matching, Rectified Flow aims to learn a linear path from source distribution $p_0(x)$ to $p(x)$\cite{RF}. The time-dependent stochastic process can be defined as
	\begin{equation}
		x_t = tx_1 + (1-t)x_0,
	\end{equation}
	where $x_0\sim p_0(x)$ and $x_1\sim p(x)$. A neural network $v_\theta(x,t)$ can be trained through
	\begin{equation}
		\mathcal{L}_{RF}=\mathbb{E}_{x_1,x_0,t}||(x_1-x_0)-v(x_t,t)||^2.
	\end{equation}
	\subsection{Network Architecture}\label{Network_Architecture}
	The backbone architecture of a generative model directly determines how its internal representations are structured and their potential for feature extraction. This section introduces the backbone networks commonly used in diffusion models and flow-based models, respectively.
	\subsubsection{Diffusion Models}
	The denoising network of diffusion models typically adopts the U-Net architecture\cite{U-Net}. A U-Net consists of an downsampling encoder and an upsampling decoder, connected by skip connections that propagate high-resolution feature maps from the encoder to the corresponding decoder layers. This structure naturally provides a spectrum of feature maps spanning from fine-grained texture in shallow layers to semantic abstraction in deep layers, which serves as a key foundation for subsequent representation extraction. The timestep $t$ is embedded into each residual block via sinusoidal positional encoding, enabling the network to perceive the current denoising stage.
	
	Latent Diffusion Models\cite{StableDiffusion} perform diffusion in the latent space of a pre-trained VAE, significantly reducing computational cost. Its U-Net backbone remains largely consistent with DDPM, but introduces an additional cross-attention mechanism to support flexible text-conditioned control. Note that the cross-attention maps proposed by this mechanism are found to naturally encode word-to-pixel semantic correspondences\cite{DAAM}.
	
	n recent years, Transformer-based backbones have also been introduced to diffusion models. The Diffusion Transformer (DiT)\cite{DiT} adopts a ViT-style architecture that partitions the input image into patches and projects them into token sequences, processed through Transformer blocks, demonstrating strong scalability.
	\begin{figure*}[t!]
		\centering
		\begin{forest}
			for tree={
				forked edges, 
				edge={-}, 
				draw=black, 
				rounded corners, 
				grow'=east, 
				l sep=8pt, 
				s sep=3pt, 
				anchor=west, 
				font=\footnotesize 
			}
			[Representation Learning in Diffusion and Flow-based Model, title, l sep=8pt
			[Generation-\\Centric\\Applications\\(Sec. \ref{Generation-Centric}), subtitle, fill=BackgroundRed
			[Generation\\(Sec. \ref{Generation}), subsubtitle, fill=BackgroundRed
			[{\textcolor{ColorA}{Diffusion}: REPA\cite{REPA}, U-REPA\cite{U-REPA}, REPA-E\cite{REPA-E}, SoftREPA\cite{SoftREPA}, VideoREPA\cite{VideoREPA}, SARA\cite{SARA},\\HASTE\cite{HASTE}, Dispersive Loss\cite{DispersiveLoss}, DiverseDiT\cite{DiverseDiT}, SRA\cite{SRA}, VA-VAE\cite{VA-VAE}, RAE\cite{RAE}, SVG\cite{SVG}, RCG\cite{RCG}}, methods, fill=BackgroundRed]
			[{\textcolor{ColorB}{Flow-based}: Self-Flow\cite{selfflow}, REPA\cite{REPA}, REG\cite{reg}, RepTok\cite{reptok}, DINO-SAE\cite{dinosae}, REPA-G\cite{repag}, $\Delta$RN\cite{RN}}, methods, fill=BackgroundRed]
			]
			[Editing\\(Sec. \ref{Editing}), subsubtitle, fill=BackgroundRed
			[{\textcolor{ColorA}{Diffusion}: Paint-by-Inpaint\cite{PaintByInpaint}, PS-Diffusion\cite{PSDiffusion}, EraDiff\cite{EraDiff}, ObjectClear\cite{ObjectClear}, YOEO\cite{YOEO},\\AnyEdit\cite{AnyEdit}, InsightEdit\cite{InsightEdit}, DragDiffusion\cite{DragDiffusion}, GeoDrag\cite{GeoDrag}, SCAdapter\cite{SCAdapter}, LightLab\cite{LightLab}, Dream-\\Light\cite{DreamLight}, IC-Light\cite{ICLight}, TextureDiffusion\cite{TextureDiffusion}, SDEdit\cite{SDEdit}, EDICT\cite{wallace2023edict}, Edit-Friendly DDPM\cite{huberman2024edit},\\Negative-Prompt Inversion\cite{NegativePromptInversion}, Direct Inversion\cite{DirectInversion}, Prompt-to-Prompt\cite{PromptToPrompt}, Plug-and-Play Diffusion\\ Features\cite{PlugAndPlayDiffusionFeatures}, MasaCtrl\cite{MasaCtrl}, KV-Edit\cite{KVEdit}, Zero-Shot I2I\cite{ZeroShotI2I}, SEGA\cite{SEGA}, UniTune\cite{UniTune}, CustomEdit\cite{CustomEdit},\\Null-Text Inversion\cite{NullTextInversion}, KV Inversion\cite{KVInversion}, DragText\cite{DragText}, DragonDiffusion\cite{DragonDiffusion}, InstructPix2Pix\cite{InstructPix2Pix},\\FireEdit\cite{FireEdit}, ControlNet\cite{ControlNet}, IP-Adapter\cite{IPAdapter}, BrushNet\cite{BrushNet}, InstructDiffusion\cite{InstructDiffusion}, MGIE\cite{MGIE}}, methods, fill=BackgroundRed]
			[{\textcolor{ColorB}{Flow-based}: OmniPaint\cite{OmniPaint}, InsertAnything\cite{InsertAnything}, Step1X-Edit\cite{Step1XEdit}, DragFlow\cite{DragFlow}, Stable Flow\cite{StableFlow},\\FlowEdit\cite{FlowEdit}, FireFlow\cite{FireFlow}, ReFlex\cite{ReFlex}, DRFS\cite{DRFS}, In-Context Edit\cite{InContextEdit}, SliderEdit\cite{SliderEdit}, Dream-\\Omni\cite{DreamOmni}, OmniGen2\cite{OmniGen2}, Qwen-Image~\cite{QwenImage}, ReasonEdit~\cite{ReasonEdit}}, methods, fill=BackgroundRed]
			]
			[Trustworthy\\Generation\\(Sec.\ref{Trustworthy Generation}), subsubtitle, fill=BackgroundRed
			[{\textcolor{ColorA}{Diffusion}: Stable Signature\cite{StableSignature}, Tree-Ring Watermark\cite{TreeRing}, Gaussian Shading\cite{GaussianShading}, RingID\cite{RingID}, PRC\\Watermark\cite{PRCWatermark}, InvisMark\cite{InvisMark}, Dynamic Watermarks\cite{DynamicWatermarks}, SynthID-Image\cite{SynthIDImage}, ShapeMark\cite{ShapeMark}, Dual-\\Guard\cite{DualGuard}, Watermarks in the Sand\cite{ImpossibilityWatermarks}, Generative Watermark Removal\cite{GenerativeWatermarkRemoval}, Vanishing Watermark-\\s\cite{VanishingWatermarks}, Crack in the Bark\cite{CrackInBark}, Warfare\cite{Warfare}, SoK\cite{SoKWatermarking}, MarkDiffusion\cite{MarkDiffusion}, BadDiffusion\cite{BadDiffusion},Eviden-\\tiary Provenance\cite{EvidentiaryProvenance}, T. Matsumoto \textit{et al.}\cite{MembershipInferenceDiffusion}, N. Carlini \textit{et al.}\cite{ExtractingTrainingData}, R. Webster\cite{ReproducibleExtraction}, Jailbreaking\\Prompt Attack\cite{JailbreakingPromptAttack}, Realistic MIA\cite{RealisticMIA}, UIBDiffusion\cite{UIBDiffusion}, Silent Branding Attack\cite{SilentBranding}, Data-Chain\\Backdoor\cite{DataChainBackdoor}, BadRSSD\cite{BadRSSD}, AdvDM\cite{AdvDM}, Glaze\cite{Glaze}, Anti-DreamBooth\cite{AntiDreamBooth}, Nightshade\cite{Nightshade},\\MetaCloak\cite{MetaCloak}, Score Distillation Protection\cite{ScoreDistillationProtection}, Mist\cite{Mist}, EditShield\cite{EditShield}, VCPro\cite{VCPro}, RID\cite{RID},\\Adv-CPG\cite{AdvCPG}, DiffusionGuard\cite{DiffusionGuard}, StyleGuard\cite{StyleGuard}, GuardDoor\cite{GuardDoor}, Anti-Diffusion\cite{AntiDiffusion}, Anti-\\Inpainting\cite{AntiInpainting}, DCT-Shield\cite{DCTShield}, WaveGuard\cite{WaveGuard}, Protection Benchmark\cite{ProtectionBenchmark}}, methods, fill=BackgroundRed]
			[{\textcolor{ColorB}{Flow-based}: Rectified Flow Tree-Ring\cite{RectifiedFlowTreeRing}, I. Baglin \textit{et al.}\cite{DeepLeakageFlowMatching}, DeContext\cite{DeContext}, Flux-Guard\cite{FluxGuard},\\Off-the-Shelf Image-to-Image Purification\cite{OffShelfPurification}, Purify Once\cite{PurifyOnce}}, methods, fill=BackgroundRed]
			]
			]
			[Perception-\\Oriented\\Applications\\(Sec. \ref{Perception-Oriented}), subtitle, fill=BackgroundYellow
			[Classification\\(Sec. \ref{Image Classification Task}), subsubtitle, fill=BackgroundYellow
			[{\textcolor{ColorA}{Diffusion}: DDAE\cite{DDAE}, GDC\cite{GDC}, DifFormer\cite{DifFormer}, DifFeed\cite{DifFormer}, A. N. Juscafresa \textit{et al.}\cite{Diffusion-Plankton}, RepFu-\\sion\cite{RepFusion}, DiDiCM\cite{DiDiCM}, ALIA\cite{ALIA}, Diff-Mix\cite{DiffMix}, AGA\cite{AGA}, DiffII\cite{DiffII}, Augmented Condition-\\ing\cite{AugmentedConditioning}, SGD-Mix\cite{SGDMix}, OntoAug\cite{OntoAug}, GenMix\cite{GenMix}, Diffusion Curriculum\cite{DiffusionCurriculum}, TADA\cite{TADA}, Stable-\\Rep\cite{StableRep}, SynCLR\cite{SynCLR}, Free-ATM\cite{Free-ATM}, DC\cite{DC}, MiPO\cite{MiPO}, MDC\cite{MDC}, Diff-Feat\cite{Diff-Feat}}, methods, fill=BackgroundYellow]
			[{\textcolor{ColorB}{Flow-based}: SymmFlow\cite{SymmFlow}, DFM\cite{DFM}} , methods, fill=BackgroundYellow]
			]
			[Dense\\Prediction\\(Sec. \ref{Dense Visual Prediction Task}), subsubtitle, fill=BackgroundYellow
			[{\textcolor{ColorA}{Diffusion}: DifFormer\cite{DifFormer}, DifFeed\cite{DifFormer}, RepFusion\cite{RepFusion}, DDPM-Seg\cite{DDPM-Seg}, DAAM\cite{DAAM}, VPD\cite{VPD},\\MDM\cite{MDM}, DDP\cite{DDP}, DMP\cite{DMP}, GenPercept\cite{GenPercept}, Dataset Diffusion\cite{DatasetDiffusion}, FreeMask\cite{FreeMask},\\ODISE\cite{ODISE}, LDIS\cite{LDIS}, ECoDepth\cite{ECoDepth}, Marigold\cite{Marigold}, Diffusion For Depth\cite{DiffusionForDepth}}, methods, fill=BackgroundYellow]
			[{\textcolor{ColorB}{Flow-based}: SemFlow\cite{SemFlow}, SymmFlow\cite{SymmFlow}, Free-ATM\cite{Free-ATM}, LawDIS\cite{LawDIS}, FlowDIS\cite{FlowDIS},\\DepthFM\cite{DepthFM}, CH3Depth\cite{CH3Depth}}, methods, fill=BackgroundYellow]
			]
			[Instance-Level\\Perception\\(Sec. \ref{Instance-Level Perception Task}), subsubtitle, fill=BackgroundYellow
			[{\textcolor{ColorA}{Diffusion}: ODISE\cite{ODISE}, VPD\cite{VPD}, LD-ZNet\cite{LD-ZNet}, DiffusionDet\cite{DiffusionDet}, DifFormer\cite{DifFormer}, DifFeed\cite{DifFormer},\\DiffusionInst\cite{DiffusionInst}, Free-ATM\cite{Free-ATM}, DALL-E Detection\cite{DALL-E-Detection}, MosaicFusion\cite{MosaicFusion}, DiffusionEngine\cite{DiffusionEngine},\\ReCon\cite{ReCon}, RepFusion\cite{RepFusion}, OC-DiT\cite{OC-DiT}, Grounded Diffusion\cite{GroundedDiffusion}}, methods, fill=BackgroundYellow]
			[{\textcolor{ColorB}{Flow-based}: RLFSeg\cite{RLFSeg}}, methods, fill=BackgroundYellow]
			]
			[Annotation-\\Scarce\\(Sec. \ref{Annotation-Scarce Task}), subsubtitle, fill=BackgroundYellow
			[{\textcolor{ColorA}{Diffusion}: DAFusion\cite{DAFusion}, DIAGen\cite{DIAGen}, DreamTeacher\cite{DreamTeacher}, iFSS-Diff\cite{iFSS-Diff}, ScribbleGen\cite{ScribbleGen},\\DreamMask\cite{DreamMask}, MaskDiff\cite{MaskDiff}, TMI\cite{TMI}, Gen-n-Val\cite{Gen-n-Val}, DiverGen\cite{DiverGen}, InstaDA\cite{InstaDA}}, methods, fill=BackgroundYellow]
			]
			]
			[Generalist\\and Unified\\Applications\\(Sec. \ref{Generalist and Unified}), subtitle, fill=BackgroundBlue
			[Clustering\\Analysis\\(Sec. \ref{Clustering Analysis}), subsubtitle, fill=BackgroundBlue
			[{\textcolor{ColorA}{Diffusion}: ClusterDDPM\cite{ClusterDDPM}, CLUDI\cite{CLUDI}, DiFiC\cite{DiFiC}, DiEC~\cite{diec}, Subspace Diffusion~\cite{diffusionsubspace}}, methods, fill=BackgroundBlue]
			[{\textcolor{ColorB}{Flow-based}: SCFM~\cite{scfm}, CPFM~\cite{cpfm}, Straight-Path FM~\cite{straightpathfm}, Latent-CFM~\cite{latentcfm}, SubFlow~\cite{subflow}}, methods, fill=BackgroundBlue]
			]
			[World Model\\(Sec. \ref{World Model}), subsubtitle, fill=BackgroundBlue
			[{\textcolor{ColorA}{Diffusion}: GameNGen\cite{valevski2024gamengen}, DIAMOND\cite{alonso2024diamond}, Cosmos\cite{nvidia2025cosmos}, JEDI\cite{lim2026jedi}, OccSora\cite{wang2024occsora}, IRASim\cite{zhu2024irasim},\\dWorldEval\cite{li2026dworldeval}}, methods, fill=BackgroundBlue]
			[{\textcolor{ColorB}{Flow-based}: GLD\cite{jang2026gld}, LaMP\cite{wang2026lamp}}, methods, fill=BackgroundBlue]
			]
			[Unified Model\\(Sec. \ref{Unified Model}), subsubtitle, fill=BackgroundBlue
			[{\textcolor{ColorA}{Diffusion}: DreamLLM\cite{dong2023dreamllm}, SEED\cite{ge2023planting}, SEED-LLaMA\cite{ge2024making}, SEED-X\cite{ge2024seed}, Emu2\cite{sun2024generative}, MM-Interleav-\\ed\cite{tian2024mm}, PUMA\cite{fang2024puma}, Unifluid\cite{fan2025unified}, BLIP3-o\cite{chen2025blip3}, OmniGen2\cite{wu2025omnigen2}, Ovis-U1\cite{wang2025ovis}, UniCode$^2$\cite{chan2025unicode2}, \\Qwen-Image\cite{wu2025qwen}, UniPic-2.0\cite{wei2025skywork}, MammothModa2\cite{shen2025mammothmoda2}, UniAR\cite{pengunified}, HyperCLOVA X\cite{team2026hyperclova}, Versati-\\leDiffusion\cite{xu2023versatile}, UniDiffuser\cite{bao2023one}, Dual Diffusion\cite{li2024dual}, UniModel\cite{zhang2025unimodel}, UniDisc\cite{swerdlow2025unified}, Lavida-O\cite{li2025lavidao},\\MMaDA\cite{yang2025mmada}, Lumina-DiMOO\cite{xin2025lumina}, Muddit\cite{shi2025muddit}, LLaDA2.0-Uni\cite{ai2026llada2}, Omni-Diffusion\cite{li2026omni},Dynin-\\Omni\cite{kim2026dynin}, ViewMask-1-to-3\cite{zhu2026viewmask}, Transfusion\cite{zhou2025transfusion}, MonoFormer\cite{zhao2024monoformer}, LMFusion\cite{shi2024llamafusion}, TUNA\cite{liu2025tuna},\\Show-o\cite{xie2024show}, UniCTokens\cite{an2026unictokens}}, methods, fill=BackgroundBlue]
			[{\textcolor{ColorB}{Flow-based}: UniCode$^2$\cite{chan2025unicode2}, Nexus-Gen\cite{zhang2025nexus}, X-Omni\cite{geng2025x}, FUDOKI\cite{wang2025fudoki}, NExT-OMNI\cite{luo2025next},\\JanusFlow\cite{ma2024janusflow}, BAGEL\cite{deng2025bagel}, Mogao\cite{liao2025mogao}, LightFusion\cite{wang2025lightfusion}, HBridge\cite{wang2025hbridge}, EMMA\cite{he2025emma}}, methods, fill=BackgroundBlue]
			]
			]
			]
		\end{forest}
		\caption{A comprehensive taxonomy of applications for diffusion models and flow-based models in representation learning. All methods are further classified according to their generative paradigm, i.e., diffusion-based versus flow-based.}
		\label{Tree}
	\end{figure*}
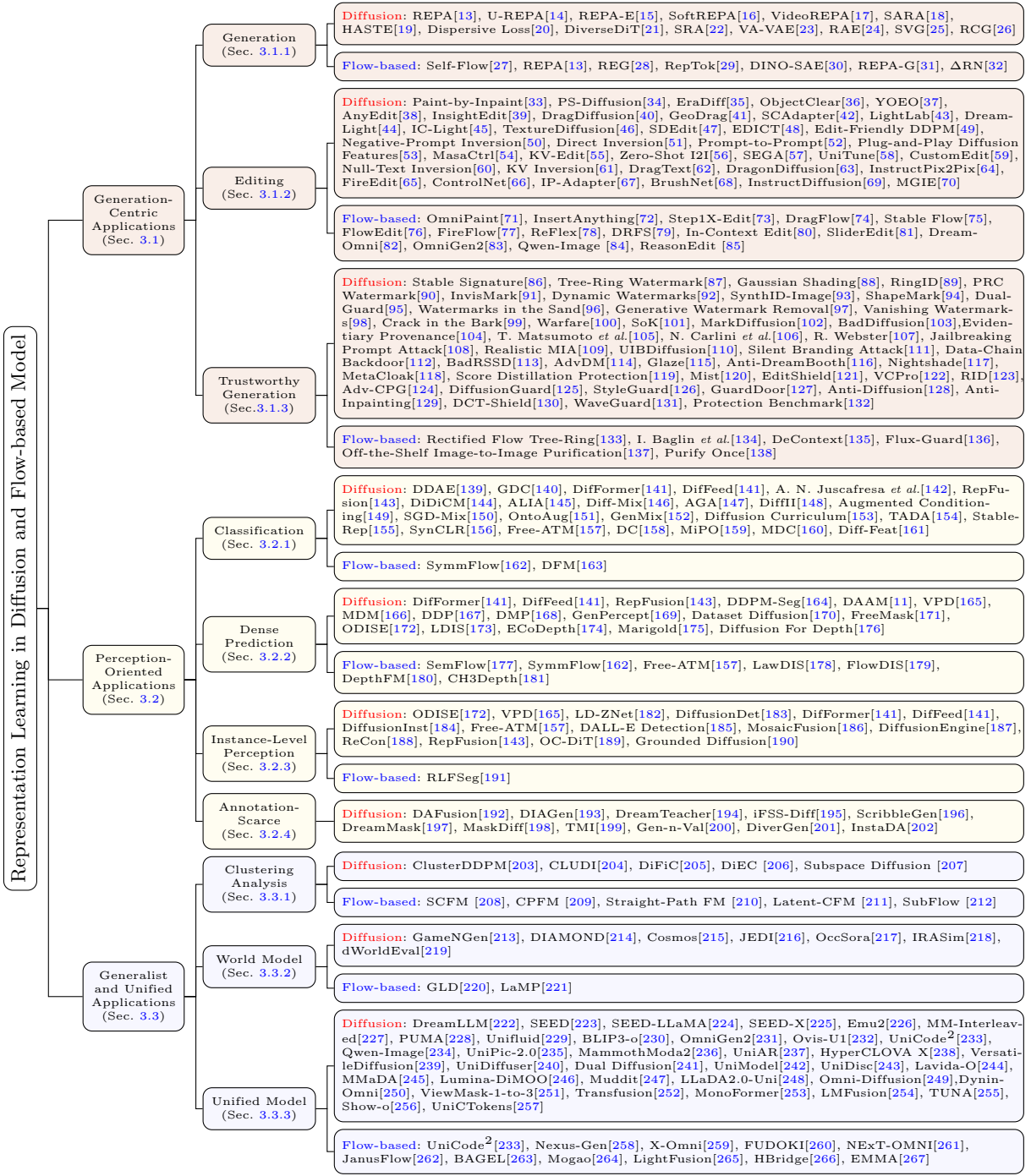
	\subsubsection{Flow-based Models}
	Flow-based models share a similar trajectory with diffusion models in backbone selection. Early works largely adopt the U-Net architecture as the parameterization network for the vector field\cite{FM}, whose encoder-decoder structure and skip connections similarly provide multi-level feature maps. As generative models evolve toward Transformer architectures, Scalable Interpolant Transformer (SiT)\cite{SiT} introduces the DiT architecture into the flow matching framework, adopting the same Transformer backbone as DiT. By learning linear interpolation paths from Gaussian noise to the data distribution, SiT demonstrates performance and scalability comparable to or even better than diffusion models on large-scale generation tasks such as ImageNet.
	
	\section{Applications}
	Building upon the background established above, this section systematically surveys the representative application scenarios of diffusion models and flow-based models in representation learning. We organize existing works into three progressive tiers, spanning from generation-centric tasks to perception-oriented tasks, and ultimately to general-purpose and unified applications. We first examine generation-centric applications, investigating how representation learning can enhance the training efficiency, generation quality, and controllability of generative models themselves, covering image and video generation, image editing, and trustworthy generation. We then turn to perception-oriented applications, focusing on how transferable representations can be extracted from pre-trained generative models to serve discriminative downstream tasks including image classification, dense visual prediction, and instance-level perception, with special attention to annotation-scarce scenarios. Finally, at the level of generalist and unified applications, we discuss how generative models transcend the traditional boundary between generation and discrimination, serving broader intelligent tasks such as clustering analysis, world model construction, and unified multimodal modeling, and summarize the corresponding architectural paradigms of unified models. All representative works mentioned in this section, along with their taxonomic relationships, are summarized in Figure \ref{Tree}.
	
	\subsection{Generation-Centric Applications}\label{Generation-Centric}
	\subsubsection{Image and Video Generating}\label{Generation}
	\paragraph{Diffusion Model}
	A large number of recent studies have validated that the generation capability of diffusion models is highly dependent on the quality of their internal representations. Whether through external knowledge injection to accelerate training, internal regularization to optimize representational structures, or fundamentally re-engineering the latent space upon which diffusion models rely, the ultimate goal is to enable generative models to "learn better and generate more faithfully." Around this core objective, this section systematically reviews representative works that leverage representations to enhance the generative capabilities of diffusion models, spanning five dimensions: external representation alignment, self-organization of internal representation, latent space paradigm upgrade, and improving generative paradigms with representations.
	
	\textbf{External Representation Alignment} is the most systematically explored approach in the direction of leveraging representations to enhance diffusion models. Its core idea is to use a pre-trained visual or language encoder as a teacher and inject discriminative representations into the intermediate layers of diffusion models, thereby compensating for the inherent semantic deficiencies of generative models. REPA\cite{REPA} is the pioneering work of this paradigm. Through simple patch-wise cosine similarity maximization, REPA aligns the intermediate representations of DiT\cite{DiT} with DINOv2\cite{DINOv2}, demonstrating for the first time that external discriminative representations can significantly accelerate diffusion model training. On ImageNet, REPA accelerates SiT training by over 17 times and achieves a state-of-the-art FID of 1.42 at the time. This discovery initiated extensive follow-up research across multiple dimensions.
	
	While REPA has been proven effective on the DiT architecture, it has not been validated on U-Net\cite{U-Net}, which exhibits faster convergence compared to DiTs. U-REPA\cite{U-REPA} fills this gap by
	proposing three adaptation strategies: mid-layer alignment, upsampling after MLP projection, and manifold loss. As a result, U-REPA achieves an FID of 1.41 at 400 epochs, surpassing 1.42 of REPA at 800 epochs.
	
	In terms of end-to-end training, REPA-E\cite{REPA-E} further back-propagates the REPA loss to the VAE while keeping the diffusion loss only updating the diffusion model, addressing the issue that direct end-to-end training with the diffusion loss causes the VAE latent space to become overly smooth. This achieves a 17$\times$ speedup over REPA and a 45$\times$ speedup over standard training, reaching an FID of 1.12.
	
	Some studies have also explored the generalizability of REPA across different modalities. SoftREPA\cite{SoftREPA} extends alignment from visual-visual to text-image modalities, leveraging contrastive learning to maximize mutual information between text and image representations. By training fewer than 1M parameters of soft tokens prepended to text features, it improves alignment quality in text-to-image generation at low cost. In terms of application scenarios, VideoREPA\cite{VideoREPA} introduces representation alignment to video generation, addressing the insufficient physical understanding of Text-to-Video models through token relation distillation loss, which aligns intra-frame spatial relationships and inter-frame temporal dynamics with Video Foundation Model (e.g. VideoMAEv2\cite{VideoMAEv2}), improving the Physical Commonsense score on VideoPhy by 24.1\% overall.
	
	As the REPA paradigm gained widespread adoption, researchers also began to reflect on its potential limitations. SARA\cite{SARA}, through singular value decomposition analysis, finds that while the patch-wise alignment is effective, it disrupts the internal structure and global distribution of the representation space. the top 50 singular values of REPA representations account for 82.6\% of the total energy, compared to only 63.9\% for DINOv2. To address this, SARA introduces structural alignment via autocorrelation matrix matching to preserve internal relational consistency and adversarial alignment through a lightweight discriminator to ensure global distribution consistency on top of patch-wise alignment, constructing a three-tier alignment framework.
	
	At the same time, HASTE\cite{HASTE}, through gradient angle analysis, reveals that the relationship between REPA gradients and diffusion loss gradients undergoes three stages. To be specific, REPA may help training in the beginning 200K iterations, the cosine similarity between$\mathcal{L}_{\text{diff}}$ and $\mathcal{L}_{\text{REPA}}$ gradually decreases to nearly orthogonal level in the following 200K iterations, indicating that it neither help nor hurt. To make matters worse, REPA may erases detail the student model tries to learn in over 400K iterations. The fundamental cause is that the capacity of the external teacher (DINOv2) is fixed, while the diffusion model requires high-frequency details in the later stages of training that the teacher cannot provide. At this point, continued alignment becomes an obstacle to performance improvement. To address this, HASTE terminating the alignment loss at an appropriate iteration to allow the diffusion model to freely develop its generative capacity. It also introduces attention alignment to distill the self-attention matrix of DINOv2 into the intermediate layers of the student model.
	
	SARA and HASTE reveal the deep contradictions of the external alignment paradigm from different angles. The former points out inherent structural defects in alignment quality, while the latter identifies that the capacity ceiling of the external teacher becomes a bottleneck for later-stage fine-tuning. These reflections drive the external alignment paradigm from whether it works toward the deeper question of how to make it work better and more sustainably. The continued evolution of the external representation alignment pathway also raises a question in another direction: \textit{Can diffusion models spontaneously organize their representational structures through internal mechanisms without external teacher dependence?} The next section discusses representative works along this direction.
	
	\textbf{Self-Organization of Internal Representation} intends to improve the quality of the representations with no external supervisory signals. Generally, it discovers and optimizes representational structures from within diffusion models themselves through carefully designed regularization losses or self-distillation mechanisms, making these structures more conducive to generation.
	
	Dispersive Loss\cite{DispersiveLoss} is an early representative of this direction. Its key insight is that the diffusion loss inherently pulls predictions toward targets. Hence, an additional loss term that pushes features from different samples away will complete the two fundamental elements of representation learning. This paper proposes a simple plug-and-play regularizer that requires no extra parameters, data augmentation, or external models. Moreover, its benefit grows with model size, as regularization may help mitigate overfitting. Under zero external dependency, it pushes the FID of SiT-XL/2 to 1.97, approaching REPA's 1.80. To be specific, the regularizer can be also applied to flow-based models.
	
	DiverseDiT\cite{DiverseDiT} goes a step further by further investigating the mechanisms underlying REPA and Dispersive Loss. Through systematic analysis of representation similarity across DiT blocks using Centered Kernel Alignment, it reveals that both REPA and Dispersive Loss essentially promote representation diversity among different blocks. Based on this insight, DiverseDiT designs long residual connections to increase input diversity for each block and a representation diversity loss to explicitly penalize feature similarity across blocks, achieving FID of 1.89 at 80 epochs and 1.52 at 200 epochs.
	
	SRA\cite{SRA} pursues a self-distillation approach, aligning the early-layer representations of the student model with the late-layer representations of an EMA teacher model. Under zero external dependency, it achieves FID of 1.58 at 800 epochs, substantially narrowing the gap to REPA with an FID of 1.42, which relies on an external model.
	
	\textbf{Latent Space Paradigm Upgrade} constitutes the third critical pathway. All the aforementioned works optimize diffusion models within a given VAE latent space. However, the weak representational capacity of the VAE itself forms a fundamental bottleneck for generation quality. Therefore, some researchers have turned to fundamentally re-engineering the latent space rather than optimizing within a given space. 
	
	VA-VAE\cite{VA-VAE} takes a moderate approach. It retains the VAE architecture but introduces guidance from Vision Foundation Models during training, using marginal cosine similarity loss and marginal distance matrix similarity loss to make the VAE latent space more uniformly distributed, achieving FID of 1.35 at 800 epochs. 
	
	RAE\cite{RAE} directly replaces the VAE encoder with a pre-trained DINOv2 encoder to produce high-dimensional semantic representations, complemented by dimension-dependent noise schedule shifting and noise-augmented decoding. It achieves FID of 1.51 (without guidance) and 1.13 (with guidance) on ImageNet 256$\times$256, with a 47$\times$ training acceleration.
	
	SVG\cite{SVG} identifies the fundamental issue of current VAE encoder. It points out that the VAE latent space lacks semantic dispersion. To address this, SVG pairs DINOv3\cite{DINOv3} with a lightweight trainable ViT residual encoder to capture fine-grained details. It is concatenated with the representations of DINOv3 after batch-level distribution alignment, enabling few-step sampling and task-general representations.
	
	These three methods progressively explore ways to improve the VAE encoder. VA-VAE optimizes the VAE within the existing framework, RAE directly replaces the VAE encoder, and SVG actively constructing a latent space that captures both semantics and fine details.
	
	\textbf{Improving Generative Paradigms with Representations} is a parallel research direction. Rather than directly modifying the internal structure or training objectives of diffusion models, it seeks to restructure the generative workflow itself, investigating how representations can reconceive the generative paradigm of diffusion models.
	
	RCG\cite{RCG}, as a representative work, aims to fundamentally address the difficulty of traditional unconditional generation in directly modeling high-dimensional pixel distributions by reformulating unconditional generation as a two-stage framework. To be specific, the model generate representations firstly, and then generate images. RCG trains a lightweight diffusion model to unconditionally generate representations in a self-supervised representation space; subsequently, these generated representations condition a standard image generator to generate images. The core advantage of this decomposition lies in the fact that the representation space is far lower in dimensionality than pixel space yet rich in high-level semantic information, making unconditionally generating representations an easier task to learn.
	
	Experiments demonstrate that RCG achieves an FID of 2.15 on ImageNet 256$\times$256, reducing the previous SOTA of 5.91 by 64\%, marking the first time that unconditional generation reaches performance comparable to conditional generation. Moreover, it brings FID improvements across various image generators (e.g. LDM, ADM, DiT, and MAGE).
	
	\paragraph{Flow-based Model}
	The representation-enhancement methods discussed in the previous section are not limited to diffusion models. External representation alignment, internal representation regularization, and representation-based latent spaces can also be applied to flow-based models such as SiT and Rectified Flow. Since these shared methods have already been introduced above, this section focuses on designs that are more closely related to continuous transport, including heterogeneous corruption, joint semantic transport, generation in representation spaces, and representation-guided sampling.
	
	\textbf{Heterogeneous Corruption and Self-Supervised Flow Learning} improves representation learning by changing how intermediate states are constructed. Standard Flow Matching usually assigns the same time step to all tokens in one sample. As a result, all regions receive a similar amount of corruption, and the model has little reason to use information from one region to understand another.
	
	Self-Flow\cite{selfflow} assigns different time steps to different token groups in the same sample. Some tokens remain relatively clean, while others are more strongly corrupted. The model must use the available context to predict the velocity of each token. This encourages it to learn semantic and structural relations between different regions.
	
	The method can be understood in two ways. First, cleaner tokens may provide useful context for predicting more corrupted tokens. Second, different token-wise time steps expose one sample to several corruption levels in a single training step. This acts as data augmentation along the time dimension. Unlike REPA-style methods, Self-Flow does not learn representations by matching an external teacher. Instead, it improves the training task used by Flow Matching itself.
	
	The same idea can be extended to video and audio generation. Different frames, regions, or temporal segments can be assigned different corruption levels, encouraging the model to learn dependencies across space and time.
	
	\textbf{Joint Transport of Visual and Semantic States} introduces semantic representations directly into the state generated by the flow model. In representation-alignment methods, semantic features are used only as training targets and disappear during inference. Joint-transport methods instead generate visual and semantic variables together.
	
	REG\cite{reg} adds a compact semantic token to the image-token sequence. During training, the semantic token is obtained from a pretrained visual encoder and paired with the image latent. The model then learns the evolution of both the image tokens and the semantic token. During sampling, both parts are initialized from noise and generated together.
	
	The semantic token therefore remains active throughout the generation process. It can provide high-level information while the image structure and visual details are being formed. This is different from REPA, where the external representation only provides additional supervision during training.
	
	Joint transport also introduces new difficulties. Visual and semantic variables may have different dimensions, distributions, and levels of uncertainty. The model must learn how these variables interact and how quickly they should evolve. Flow Matching makes it possible to use different time schedules for the two parts. For example, semantic information may be generated earlier and then guide the later formation of image details.
	
	\textbf{Flow Matching in Representation-Native Latent Spaces} changes the space in which the transport process is learned. The previous section has already discussed the general idea of replacing VAE latents with features from pretrained visual encoders. For Flow Matching, the main question is whether the interpolation path is suitable for the geometry of these features.
	
	RAE can be combined directly with Flow Matching. A frozen visual encoder maps images into a semantic feature space, a decoder reconstructs images from these features, and a flow model learns to generate the feature distribution. Compared with conventional VAE latents, these representations contain stronger semantic information. However, they are often high-dimensional, strongly correlated, and different from a simple Gaussian distribution. Their feature scale, noise schedule, and model width therefore need to be carefully designed.
	
	RepTok~\cite{reptok} uses a small number of continuous semantic tokens instead of a dense spatial latent grid. This reduces the sequence length and lowers the cost of Flow Matching training. The main challenge is information preservation. A compact token set must retain both high-level semantics and the spatial details needed for image reconstruction.
	
	DINO-SAE~\cite{dinosae} further considers the geometry of pretrained representations. Normalized DINO features are close to a spherical space, while standard Flow Matching usually assumes Euclidean interpolation. DINO-SAE therefore combines a spherical autoencoder with Riemannian Flow Matching, so that the transport path follows the geometry of the representation space. This avoids moving through regions that are far from the valid feature distribution.
	
	\textbf{Representation-Guided ODE Sampling} uses representations to control the trajectory of a trained flow model. Since Flow Matching generates samples by solving an ODE, an additional representation objective can be used to adjust the predicted velocity during sampling.
	
	REPA-G~\cite{repag} uses the aligned feature space of a pretrained flow model for inference-time guidance. A reference image, a selected image region, or several visual concepts are first converted into target features. During sampling, the flow trajectory is adjusted so that the generated representation becomes closer to these targets.
	
	Global representations can control the overall content of an image, while patch-level representations can control local regions or visual details. Several target representations can also be combined for compositional generation. Unlike training-time alignment, this approach directly changes the sampling trajectory and therefore requires no additional model training.
	
	The guidance strength usually depends on time. At the beginning of sampling, the state is dominated by noise and does not contain stable semantic information. Near the end, strong guidance may disturb texture and fine details. Representation guidance is therefore more suitable for the middle stage, when the global structure has started to appear but can still be adjusted.
	
	\textbf{Positive-incentive noise for flow-based models} is an emerging direction that enhances flow-based generative models by rethinking the role of stochasticity. The concept of Positive-incentive Noise ($\pi$-noise)\cite{PiNoise} formalizes this idea by identifying noise signals that maximize the mutual information between the task objective and the injected randomness, thereby transforming noise from a mere disturbance into a beneficial driving force. Beyond generative modeling, the ?-noise principle has demonstrated broad potential in supervised learning\cite{VPN}, representation learning\cite{PiNDA}, graph representation learning\cite{PiNGDA}, class incremental learning\cite{MiN}, and multimodal models\cite{PiNI, MuNG}, underscoring its general utility in machine learning. For flow-based models, this principle is instantiated as Rectified Noise\cite{RN}, which introduces a lightweight learnable noise generator atop a pre-trained rectified flow model to produce $\pi$-noise conditioned on representations and inject it into the velocity field. This incurs minimal parameter overhead while yielding non-trivial performance gains, reducing the FID on ImageNet-1K from 10.16 to 9.05 with only 0.39\% extra parameters.
	
	\begin{table*}[t!]
		\centering
		\caption{Taxonomy of image editing methods.}
		\label{tab:image-editing-taxonomy}
		\begin{tblr}{
				width=\textwidth,
				colspec={
					Q[c,m,wd=0.1\textwidth]
					Q[c,m,wd=0.1\textwidth]
					Q[c,m,wd=0.2\textwidth]
					X[l,m]
				},
				row{1}={font=\bfseries},
				cell{1}{1-4}={c},
			}
			\toprule
			Taxonomy Perspective
			&
			Category
			&
			Editing Type
			&
			Representative References \\
			\midrule
			\SetCell[r=6]{c,m} Visual target perspective
			& \SetCell[r=2]{c,m} Content-level
			& Object editing
			& Paint-by-Inpaint~\cite{PaintByInpaint}, OmniPaint~\cite{OmniPaint}, InsertAnything~\cite{InsertAnything}, PS-Diffusion~\cite{PSDiffusion}, EraDiff~\cite{EraDiff}, ObjectClear~\cite{ObjectClear}, YOEO~\cite{YOEO} \\
			&
			& Scene-context editing
			& AnyEdit~\cite{AnyEdit}, Step1X-Edit~\cite{Step1XEdit}, InsightEdit~\cite{InsightEdit} \\
			\cmidrule{2-4}
			&
			\SetCell[r=4]{c,m} Expression-level
			& Structure editing
			& DragGAN~\cite{DragGAN}, DragDiffusion~\cite{DragDiffusion}, GeoDrag~\cite{GeoDrag}, DragFlow~\cite{DragFlow} \\
			&
			& Style editing
			& SCAdapter~\cite{SCAdapter} \\
			&
			& Lighting editing
			& LightLab~\cite{LightLab}, DreamLight~\cite{DreamLight}, IC-Light~\cite{ICLight} \\
			&
			& Texture editing
			& TextureDiffusion~\cite{TextureDiffusion} \\
			\midrule
			\SetCell[r=12]{c,m} Model adaptation perspective
			& \SetCell[r=5]{c,m} Training-free editing
			& Perturb-and-denoise editing
			& SDEdit~\cite{SDEdit} \\
			&
			& Inversion-based editing
			& EDICT~\cite{wallace2023edict}, Edit-Friendly DDPM~\cite{huberman2024edit}, Negative-Prompt Inversion~\cite{NegativePromptInversion}, Direct Inversion~\cite{DirectInversion} \\
			&
			& Attention-feature editing
			& Prompt-to-Prompt~\cite{PromptToPrompt}, Plug-and-Play~\cite{PlugAndPlayDiffusionFeatures}, MasaCtrl~\cite{MasaCtrl}, Stable Flow~\cite{StableFlow}, KV-Edit~\cite{KVEdit} \\
			&
			& Semantic-direction editing
			& Zero-Shot I2I~\cite{ZeroShotI2I}, SEGA~\cite{SEGA} \\
			&
			& Flow-based editing
			& FlowEdit~\cite{FlowEdit}, FireFlow~\cite{FireFlow}, ReFlex~\cite{ReFlex}, DRFS~\cite{DRFS} \\
			\cmidrule{2-4}
			&
			\SetCell[r=3]{c,m} Test-time adaptation
			& Model-parameter adaptation
			& UniTune~\cite{UniTune}, CustomEdit~\cite{CustomEdit} \\
			&
			& Embedding adaptation
			& Null-Text Inversion~\cite{NullTextInversion}, KV Inversion~\cite{KVInversion}, DragText~\cite{DragText} \\
			&
			& Drag-based adaptation
			& DragDiffusion~\cite{DragDiffusion}, DragonDiffusion~\cite{DragonDiffusion}, DragFlow~\cite{DragFlow}, GeoDrag~\cite{GeoDrag} \\
			\cmidrule{2-4}
			&
			\SetCell[r=4]{c,m} Training-based editing
			& Full-model editing
			& InstructPix2Pix~\cite{InstructPix2Pix}, AnyEdit~\cite{AnyEdit}, FireEdit~\cite{FireEdit}, InsightEdit~\cite{InsightEdit}, Step1X-Edit~\cite{Step1XEdit} \\
			&
			& Parameter-efficient editing
			& ControlNet~\cite{ControlNet}, IP-Adapter~\cite{IPAdapter}, BrushNet~\cite{BrushNet}, In-Context Edit~\cite{InContextEdit}, SliderEdit~\cite{SliderEdit} \\
			&
			& Unified-architecture editing
			& InstructDiffusion~\cite{InstructDiffusion}, MGIE~\cite{MGIE}, OmniPaint~\cite{OmniPaint}, DreamOmni~\cite{DreamOmni}, OmniGen2~\cite{OmniGen2}, Qwen-Image~\cite{QwenImage} \\
			&
			& Reasoning-enhanced editing
			& ReasonEdit~\cite{ReasonEdit} \\
			\bottomrule\hline
		\end{tblr}
	\end{table*}
	
	\subsubsection{Image Editing}\label{Editing}
	Image editing aims to modify an existing image according to a user intention while preserving the irrelevant content in the source image.
	Different from pure generation, editing requires the model to synthesize plausible pixels while deciding which factors to change or preserve.
	Image editing covers a wide range of tasks, such as object insertion or removal, background replacement, structure manipulation, style transfer and multi-reference composition.
	To systematically review the image editing field, we organize existing works from two perspectives: visual targets and model adaptation.
	Table \ref{tab:image-editing-taxonomy} provides a comprehensive taxonomy of representative image editing methods according to these two perspectives.

	\paragraph{Visual target perspective}
	According to the edited visual component, image editing works can be divided into \textbf{content-level editing} and \textbf{expression-level editing}.
	Content-level editing manipulates semantic entities and scene context, while expression-level editing modifies visual attributes such as structure, style, lighting, and texture.
	
	\textbf{Content-level methods} can be further discussed by the edited semantic component, mainly including object editing and scene-context editing.
	The core challenge of content-level editing is to localize the intended semantic change and reduce collateral effects.
	\textbf{Object editing} covers adding, removing, or replacing foreground entities.
	Adding-oriented methods address object--background composition, inpainting-based synthesis, or reference-guided appearance transfer so that the inserted entity matches the intended object and fits the surrounding scene~\cite{PaintByInpaint,OmniPaint,InsertAnything,PSDiffusion}.
	Removing-oriented methods instead emphasize accurate localization, object-effect suppression, and background completion, where the main challenge is to erase the selected entity without leaving structural artifacts, shadows, or inconsistent textures~\cite{EraDiff,ObjectClear,YOEO}.
	Replacing-oriented methods combine removal and insertion, so the edited result must align the target region layout with the new object's identity, scale, pose, and local interactions with the scene~\cite{OmniPaint,InsertAnything,PSDiffusion}.
	\textbf{Scene-context editing} is mainly integrated into general-purpose editing models rather than handled by a background-specific model.
	Multi-task and instruction-following editing models extend editable content from foreground objects to broader scene context~\cite{AnyEdit,Step1XEdit,InsightEdit}.
	
	\textbf{Expression-level methods} can be organized by the edited visual attribute, including structure, style, lighting, and texture.
	Different from content-level editing, expression-level editing keeps the main semantic entities recognizable while changing their geometric arrangement or visual appearance.
	\textbf{Structure editing} modifies pose, shape, layout, or spatial correspondence.
	Drag-based structure editing includes point-based local manipulation and region-based geometric control.
	Point-based methods propagate sparse handle-point displacements to local deformations~\cite{DragGAN,DragDiffusion,GeoDrag}, whereas DragFlow exploits FLUX's DiT prior and region-based affine supervision for relocation, deformation, and rotation~\cite{DragFlow}.
	Beyond direct source-image editing, layout-conditioned generation provides another form of structural control by specifying object locations, instance layouts, or grounded regions before image synthesis.
	GLIGEN, MIGC, and Laytrol synthesize images from grounded or instance-level layouts rather than modifying an existing image, and are therefore reviewed as generation-side structural control methods~\cite{GLIGEN,MIGC,Laytrol}.
	\textbf{Style editing} transfers appearance statistics or reference style while balancing stylization strength with semantic recognizability~\cite{SCAdapter}.
	StyleAligned instead targets style-consistent generation across multiple synthesized images and is better categorized as style-aligned generation~\cite{StyleAligned}.
	\textbf{Lighting editing} realizes illumination control through light-source adjustment, relighting harmonization, or light-transport consistency, with the key goal of aligning foreground and background illumination~\cite{LightLab,DreamLight,ICLight}.
	\textbf{Texture editing} refines fine-grained surface appearance, where the main difficulty lies in separating material-level details from object semantics~\cite{TextureDiffusion}.
	
	\paragraph{Model adaptation perspective}
	From the perspective of model adaptation, different strategies balance editing effectiveness and computational burden in different ways.
	Existing methods can be broadly divided into training-free editing, test-time adaptation, and training-based editing.
	
	\textbf{Training-free editing} keeps the pretrained generative model frozen and changes the inference process directly.
	Training-free editing includes perturb-and-denoise methods, inversion-based methods, and internal-representation intervention.
	SDEdit adds noise to the input image and denoises it under a target condition without explicit inversion~\cite{SDEdit}.
	Inversion-based methods instead reconstruct source trajectories or latent states before editing~\cite{NegativePromptInversion,DirectInversion}.
	The key challenge is to balance input fidelity with sufficient generative flexibility to achieve the desired modification.
	Attention- and feature-based methods manipulate attention maps or intermediate features to reuse source-image structure while changing the semantic target~\cite{PromptToPrompt,PlugAndPlayDiffusionFeatures,MasaCtrl,StableFlow,KVEdit}.
	Compared with sampling-only methods, attention- and feature-based methods offer finer control over identity, layout, and background preservation, but performance becomes sensitive to the selected internal representations and intervention strength.
	Text-condition and semantic-direction methods steer editing by constructing or manipulating semantic directions in the conditioning space~\cite{ZeroShotI2I,SEGA}.
	As generative backbones move from U-Net-based diffusion models to DiT-based flow-matching models, recent editing methods increasingly exploit rectified-flow trajectories, inversion processes, intermediate features, velocity updates, or optimization within flow-based backbones~\cite{FlowEdit,FireFlow,ReFlex,StableFlow,DragFlow,DRFS}.
	Training-free editing is therefore attractive for flexibility and low training cost, but the absence of learned task adaptation makes the result depend heavily on inversion strength, feature selection, and guidance scale.
	
	\textbf{Test-time adaptation} replaces large-scale editing training with instance-level adaptation during inference.
	Model-parameter customization fine-tunes the generator on a single image or a few references to capture instance-specific appearance~\cite{UniTune,CustomEdit}.
	Embedding optimization keeps the generator weights fixed and optimizes unconditional, KV, or text embeddings within an editable semantic space~\cite{NullTextInversion,KVInversion,DragText}.
	Optimization-based drag methods iteratively update latent or feature representations under point constraints~\cite{DragDiffusion,DragonDiffusion}.
	GeoDrag performs geometry-guided editing in a single forward pass, separating it from iterative drag optimization~\cite{GeoDrag}.
	Compared with training-free editing, test-time adaptation better fits the current input, but the additional optimization introduces slower inference and the risk that strong fitting may restrict large semantic changes.
	
	\textbf{Training-based editing} learns generalizable editing behavior through offline optimization on paired editing data or large-scale multi-task data.
	At the backbone level, full or substantial model adaptation directly internalizes the conditional mapping from a source image and an editing instruction to the desired output.
	Such methods typically avoid input-specific optimization at inference time, but editing coverage, instruction fidelity, and region accuracy remain strongly dependent on the diversity and alignment of the training data~\cite{InstructPix2Pix,AnyEdit,FireEdit,InsightEdit,Step1XEdit}.
	Parameter-efficient approaches instead keep most of the pretrained generator frozen and introduce trainable control branches, visual adapters, inpainting modules, or low-rank adaptation layers~\cite{ControlNet,IPAdapter,BrushNet,InContextEdit,SliderEdit}.
	While ControlNet and IP-Adapter provide general conditioning mechanisms that can be reused for editing, editing-oriented modules such as BrushNet, InContextEdit, and SliderEdit learn more specialized ways of injecting spatial, visual, or continuous-control signals.
	These approaches reduce the number of trainable parameters and preserve the pretrained generative prior, although the auxiliary modules must still avoid overwriting source identity, spatial structure, and unmodified regions.
	Beyond the location of trainable parameters, dedicated and unified architectures reorganize the editing interface itself.
	Dedicated systems jointly model tightly coupled editing processes or combine multimodal instruction understanding with image generation, whereas unified models share representations across generation, editing, reference-conditioned synthesis, and other visual tasks~\cite{InstructDiffusion,MGIE,OmniPaint,DreamOmni,OmniGen2,QwenImage}.
	Recent reasoning-enhanced frameworks further augment such editing models with explicit instruction interpretation, result reflection, and iterative correction~\cite{ReasonEdit}.
	Overall, the central challenge of training-based editing is to learn representations that generalize across diverse transformations while keeping semantic identity, spatial structure, and visual appearance sufficiently disentangled.
	
	\subsubsection{Trustworthy Generation: Watermarking, Backdoors, and Adversarial Protection}\label{Trustworthy Generation}
	
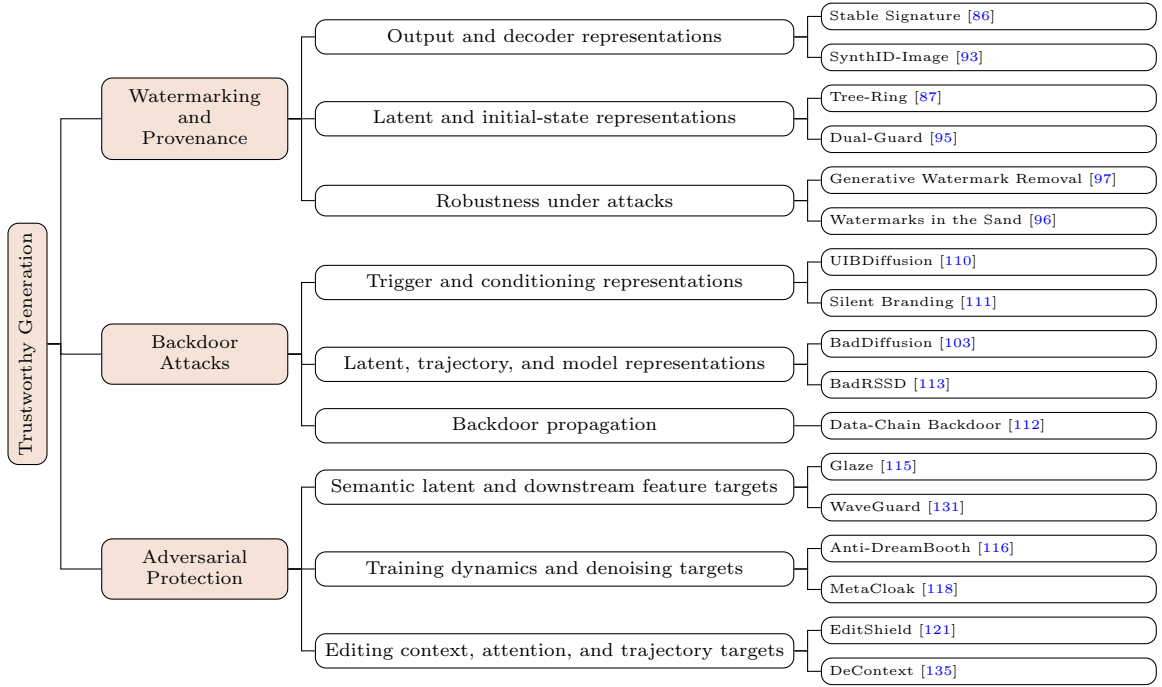
\begin{figure*}[t]
		\centering
		\begin{forest}
			for tree={
				forked edges,
				edge={-},
				draw=black,
				rounded corners,
				grow'=east,
				l sep=10pt,
				s sep=5pt,
				anchor=west,
				font=\footnotesize
			}
			[Trustworthy Generation, figuretitle, fill=BackgroundRed
			[Watermarking\\and\\Provenance, figurebranch, fill=BackgroundRed
			[Output and decoder representations, figureleaf
			[Stable Signature \cite{StableSignature}, figuremethod]
			[SynthID-Image \cite{SynthIDImage}, figuremethod]
			]
			[Latent and initial-state representations, figureleaf
			[Tree-Ring \cite{TreeRing}, figuremethod]
			[Dual-Guard \cite{DualGuard}, figuremethod]
			]
			[Robustness under attacks, figureleaf
			[Generative Watermark Removal \cite{GenerativeWatermarkRemoval}, figuremethod]
			[Watermarks in the Sand \cite{ImpossibilityWatermarks}, figuremethod]
			]
			]
			[Backdoor\\Attacks, figurebranch, fill=BackgroundRed
			[Trigger and conditioning representations, figureleaf
			[UIBDiffusion \cite{UIBDiffusion}, figuremethod]
			[Silent Branding \cite{SilentBranding}, figuremethod]
			]
			[{Latent, trajectory, and model representations}, figureleaf
			[BadDiffusion \cite{BadDiffusion}, figuremethod]
			[BadRSSD \cite{BadRSSD}, figuremethod]
			]
			[Backdoor propagation, figureleaf
			[Data-Chain Backdoor \cite{DataChainBackdoor}, figuremethod]
			]
			]
			[Adversarial\\Protection, figurebranch, fill=BackgroundRed
			[Semantic latent and downstream feature targets, figureleaf
			[Glaze \cite{Glaze}, figuremethod]
			[WaveGuard \cite{WaveGuard}, figuremethod]
			]
			[Training dynamics and denoising targets, figureleaf
			[Anti-DreamBooth \cite{AntiDreamBooth}, figuremethod]
			[MetaCloak \cite{MetaCloak}, figuremethod]
			]
			[{Editing context, attention, and trajectory targets}, figureleaf
			[EditShield \cite{EditShield}, figuremethod]
			[DeContext \cite{DeContext}, figuremethod]
			]
			]
			]
		\end{forest}
		\caption{Task-first and representation-centered organization of trustworthy generation. Each task is structured by the functional role of representations: watermark carriers, backdoor injection or storage sites, and downstream targets of adversarial protection. Each representation category is illustrated with one or two representative works; the references are not intended to be exhaustive. Purification and model mismatch are treated as cross-cutting evaluation challenges rather than standalone representation categories.}
		\label{fig:trustworthy-generation}
	\end{figure*}

	Generation-centric applications are not limited to producing and editing visual content. Learned representations can also encode information that provides evidence of content provenance. Meanwhile, backdoors can be injected by modifying internal representations of the generator, whereas adversarial protection methods manipulate selected representations to protect released visual assets from unauthorized training or editing. We therefore organize trustworthy generation around three application tasks: watermarking, backdoor injection, and adversarial protection. For each task, we examine which representation levels are involved and how they function in achieving the corresponding application objective. Figure \ref{fig:trustworthy-generation} and Table \ref{tab:trustworthy-taxonomy} present the overall taxonomy of these tasks from a representation-centered perspective, structured by the functional role that representations play across watermarking, backdoor attacks, and adversarial protection.
	
	\paragraph{Scope and task-specific representation taxonomy}
	The generation lifecycle progresses from conditioning and initial latent states, through denoising or flow trajectories, to generated outputs that may later be distributed, edited, or reused for downstream training. Throughout this lifecycle, different representations can serve as functional components for trustworthy generation tasks. The same representation may participate in multiple tasks, but its functional role varies across applications. In watermarking, it functions as a \textit{carrier} for encoding provenance information. In backdoor attacks, it provides a \textit{medium} for embedding trigger-related features that can be activated during inference. In adversarial protection, it serves as an \textit{intervention target} where perturbations are introduced to disrupt downstream learning or inference. Each discussion is organized around a common structure: application task, representation, intervention or attack stage, mechanism, evaluation, diffusion?flow differences, and limitations.

	\begin{table*}[t!]
		\centering
		\caption{Task-specific roles of representations in trustworthy generation. Representative references are illustrative rather than exhaustive.}
		\label{tab:trustworthy-taxonomy}
		\small
		\renewcommand{\tabularxcolumn}[1]{m{#1}}
		\begin{tabularx}{\textwidth}{>{\centering\arraybackslash}m{0.16\textwidth} >{\raggedright\arraybackslash}m{0.26\textwidth} >{\raggedright\arraybackslash}m{0.26\textwidth} >{\raggedright\arraybackslash}X}
			\toprule
			\multicolumn{1}{>{\centering\arraybackslash}m{0.14\textwidth}}{\textbf{Task}} &
			\multicolumn{1}{>{\centering\arraybackslash}m{0.28\textwidth}}{\textbf{Role of Representation}} &
			\multicolumn{1}{>{\centering\arraybackslash}m{0.26\textwidth}}{\textbf{Main Representation Levels}} &
			\multicolumn{1}{>{\centering\arraybackslash}X}{\textbf{Representative References}} \\
			\midrule
			Watermarking 
			& A carrier whose identity signal must propagate to the output and remain recoverable.
			& Output/decoder, initial noise or latent, final latent, and inversion trajectory.
			& Stable Signature \cite{StableSignature}; Tree-Ring \cite{TreeRing}; Gaussian Shading \cite{GaussianShading}\\
			\midrule
			
			Backdoor attacks 
			& A representation in which a trigger, semantic association or synthetic-data signal is injected or modified.
			& Data poisoning, model training or fine-tuning, triggered inference, and downstream synthetic-data reuse.
			& UIBDiffusion \cite{UIBDiffusion}; BadDiffusion \cite{BadDiffusion}; Data-Chain Backdoor \cite{DataChainBackdoor} \\
			\midrule
			
			\makecell[c]{Adversarial protection} 
			& A target representation used by personalization, editing, or student-model training.
			& Semantic/identity features, encoder latents, score or velocity objectives, attention/trajectory states, and released outputs.
			& AdvDM \cite{AdvDM}; Anti-DreamBooth \cite{AntiDreamBooth}; EditShield \cite{EditShield} \\
			\bottomrule
		\end{tabularx}
	\end{table*}
	
	\paragraph{Task I: Watermarking and provenance}
	Image watermarking aims to attach origin, attribution, or integrity evidence to generated content. It supports provenance verification, ownership attribution, and content integrity assessment. Traditional watermarking methods typically process images with dedicated watermark encoders and decoders to embed and recover information. With the development of generative models, watermarking methods have gradually shifted from post-processing the generated image to integrating watermark signals into the generation process. Existing in-generation watermarking approaches either modify the generator to produce inherently watermarked outputs or encode watermark information into generation-related representations such as latent states and initial noise. Therefore, the central representation question is where watermark information is carried and how this carrier can be reliably recovered.
	
	\representationheading{Output and decoder representations}
	
	At the output and decoder representation level, the goal is to make the released image itself carry recoverable provenance evidence, in a way that remains close to traditional image watermarking. Here, the carrier is the output-space representation, and the main design question is how to embed provenance signals into the visible image while preserving image quality and robustness. Early work such as Stable Signature\cite{StableSignature} moves watermarking into the generative pipeline by fine-tuning the latent decoder, so that provenance is rooted in the model-specific decoding process rather than added as a purely external post-processing signal. Later methods shift toward more flexible and deployable output-level watermarking. InvisMark\cite{InvisMark} and SynthID-Image\cite{SynthIDImage} adopt encoder--decoder frameworks that attach imperceptible signals to generated images after synthesis, emphasizing high-resolution deployment, robustness under common transformations, and practical provenance verification at scale. Dynamic Watermarks\cite{DynamicWatermarks} further extends this line by combining a fixed model-level watermark with content-adaptive watermark components, reflecting a transition from static source attribution to more fine-grained and image-specific traceability. Overall, the evolution at this representation level moves from decoder-integrated model attribution to scalable post-hoc provenance marking, and then to hybrid designs that jointly pursue persistence, adaptability, and deployment efficiency.
	
	\representationheading{Latent and initial-state representations}
	
	The second line of work moves the watermark carrier from the released image to the generation process itself. The motivation is that output-space signals are directly exposed after release, whereas watermark information embedded in latent or initial-state representations can be propagated through sampling and may better preserve visual fidelity. Tree-Ring\cite{TreeRing} establishes this direction by encoding a structured pattern in the Fourier representation of the initial Gaussian noise, so that provenance is tied to the sampling state rather than appended to the final image. This design makes the watermark diffusion-native, but also makes recovery depend on inversion quality. Subsequent methods refine this idea toward different objectives. Gaussian Shading\cite{GaussianShading} seeks to preserve the Gaussian latent distribution while encoding watermark information, aiming to reduce the performance loss and statistical distortion introduced by earlier noise-based schemes. RingID\cite{RingID} further extends latent watermarking from binary verification to multi-key identification, motivated by the need for scalable source tracing and user-level attribution. PRC Watermark\cite{PRCWatermark} pursues stronger security by using pseudorandom error-correcting codes to hide the watermark. ShapeMark\cite{ShapeMark} revisits the representation granularity of latent watermarking itself. Instead of encoding bits in individual noise values, it uses structural encoding through group-level shapes and permutations, with the explicit goal of improving robustness without sacrificing generation diversity. More recently, Dual-Guard\cite{DualGuard} shows that latent watermarking can also move beyond provenance-only verification by combining an initial-noise watermark for source authentication with a final-latent fingerprint for tamper localization. Overall, the development at this representation level proceeds from diffusion-native provenance embedding to distribution-preserving and multi-key designs for scalable identification. The main limitations remain inversion accuracy, scheduler or solver dependence, key security, and the difficulty of transferring these designs from diffusion models to flow-based generators where initial-state recovery follows a different trajectory formulation and numerical path\cite{RectifiedFlowTreeRing}.
	
	\representationheading{Watermark robustness under attacks}
	
	Once watermark information is embedded into either outputs or generation states, the central question is no longer only where the watermark is carried, but whether that carrier remains reliable under adversarial manipulation. This shifts the focus from watermark design to watermark survival. For output-level carriers, a growing body of work shows that imperceptible pixel-space signals can be removed by regeneration or editing while preserving semantic content. Generative reconstruction attacks\cite{GenerativeWatermarkRemoval} demonstrate that diffusion-based restoration can erase conventional invisible watermarks more effectively than destructive perturbations. Vanishing Watermarks\cite{VanishingWatermarks} further shows that diffusion editing can drive watermark information toward near-zero recoverability. These results motivate the move from exposed output signals to generation-aware carriers. For latent carriers, however, robustness is not guaranteed either. A Crack in the Bark\cite{CrackInBark} shows that public generative components such as VAEs can be exploited to approximate latent spaces, train surrogate detectors, and attack Tree-Ring-style watermarking without full model access. Warfare\cite{Warfare} broadens the threat model further by considering both removal and forgery, emphasizing that provenance systems must prevent not only false negatives but also false attribution. At a more fundamental level, Watermarks in the Sand\cite{ImpossibilityWatermarks} formalizes the quality--robustness tension and shows that strong robustness guarantees may be unattainable under sufficiently capable attackers. Recent surveys and toolkits such as SoK Watermarking and MarkDiffusion\cite{SoKWatermarking,MarkDiffusion} therefore frame watermark robustness as a threat-model-dependent property rather than a single benchmark number. Taken together, these studies show that output and latent representations fail in different ways: output carriers are directly exposed to regeneration and editing, whereas latent carriers depend on inversion secrecy, public component access, and model-specific recovery assumptions. This comparison makes representation choice central to watermark design, because robustness depends not only on the embedding algorithm but also on which representation is trusted to retain provenance under adaptive attacks.

	\paragraph{Task II: Backdoor attacks on generative models}
	Backdoor attacks study how a compromised generator can behave normally on clean inputs while producing an attacker-specified output once a hidden condition is satisfied. For this safety problem, the key representation question is not where malicious behavior is merely observed, but where the malicious association is injected, stored, and later activated. In most cases, the intervention occurs during poisoned-data construction, model training, or fine-tuning, whereas activation happens during conditional generation or downstream reuse.
	
	\representationheading{Trigger and conditioning representations}
	
	One line of work injects the backdoor at the trigger or conditioning level by associating an imperceptible or hidden condition with malicious generation behavior. Here, the injected representation is not the final output itself, but the input-side signal that steers the model toward an attacker-chosen target. UIBDiffusion\cite{UIBDiffusion} represents this direction by introducing a universal imperceptible image trigger through poisoned training data, so that the trigger becomes a reusable conditioning cue across different inputs and models. In this case, the malicious association is written into trigger-conditioned generation through pixel-space perturbations that remain visually inconspicuous. Silent Branding\cite{SilentBranding}, in contrast, removes the need for an explicit trigger and instead poisons the training set with repeated visual branding patterns. This shifts the injected representation from an explicit trigger signal to a learned semantic association between recurring visual patterns and normal generation conditions.
	
	These methods are typically evaluated in terms of attack success rate, trigger invisibility or stealth, clean-generation quality, and target specificity. Trigger and conditioning attacks are conceptually portable across diffusion and flow-based generators because both rely on conditional representations to guide generation. However, the exact strength and persistence of the learned association still depend on the underlying training objective, architecture, and conditioning interface.
	
	\representationheading{Latent, trajectory, and model representations}
	
	A second line of work moves the backdoor from the input condition to the generator's internal representations and transition dynamics. At this level, the malicious behavior is stored in the model parameters, latent semantics, or denoising trajectory, rather than solely in an external trigger pattern. BadDiffusion\cite{BadDiffusion} is the representative starting point of this direction: it modifies training or fine-tuning so that the presence of a trigger redirects the denoising process toward an attacker-selected target. The backdoor is therefore embedded in the learned reverse process itself. BadRSSD\cite{BadRSSD} further deepens this idea by aligning poisoned semantic latents with a target image and coordinating constraints across latent, pixel, feature, and trajectory spaces. Compared with simple trigger-based poisoning, this design makes the backdoor more distributed across the representation space and less dependent on a single visible trigger location. The main attack stage here is model training or adaptation, and the core mechanism is to alter the generator so that a specific initial state, semantic condition, or hidden feature configuration follows a malicious generative path. This line of work develops from trigger-conditioned trajectory manipulation to more distributed backdoors embedded in latent semantics and internal transition dynamics, which also makes detection and auditing more difficult.

	\representationheading{Backdoor propagation and open limitations}
	
	A related extension asks whether a backdoor can survive beyond the original compromised generator. Data-Chain Backdoor\cite{DataChainBackdoor} shows that generated samples themselves can act as carriers of malicious behavior: a poisoned diffusion model first produces synthetic data containing hidden triggers, and the malicious association is then inherited by a downstream model trained on those outputs. In this setting, the intervention occurs during synthetic-data generation and downstream training rather than only within the original sampler. This broadens the representation question from where the backdoor is embedded inside a single generator to how malicious associations propagate through the generative data supply chain. However, current evidence for such propagation remains heavily diffusion-centered. It is still unclear whether conclusions drawn from diffusion models transfer directly to flow-based generators, or whether flow-specific properties such as velocity prediction and continuous-time trajectories change the propagation, persistence, or detectability of backdoor behavior.

	\paragraph{Task III: Adversarial protection against unauthorized reuse}
	Adversarial protection aims to prevent released images or generated outputs from supporting unauthorized personalization, style imitation, editing, or knowledge distillation. For this task, the central representation question is which downstream representation or learning signal is disrupted. In most settings, the released perturbation is only the carrier, while the actual target is the latent feature, semantic embedding, denoising signal, attention pathway, or reconstructed representation that an unauthorized reuser depends on. In this section, we first discuss two anti-training levels of protection, namely semantic-feature attacks and training-dynamics attacks, and then turn to the task of anti-editing.
	
	\representationheading{Semantic latent and downstream feature representations}
	
	A first line of adversarial protection targets latent semantic features that support style imitation, identity personalization, concept extraction, or student-side feature learning. The goal is to prevent downstream models from recovering protected semantics while keeping the released image visually usable. Here, the attacked representations include style embeddings, identity features, concept-level semantic spaces, and latent features extracted from released outputs. AdvDM\cite{AdvDM} initiates this direction by disrupting the feature extraction process used in painting imitation. Glaze\cite{Glaze} similarly shifts style-related representations so that models trained on protected artworks learn misleading style cues. Mist\cite{Mist} strengthens this line by jointly perturbing semantic and texture-related signals to improve transfer across personalization pipelines. VCPro\cite{VCPro} further refines the objective by selectively protecting owner-specified concepts or regions while preserving non-target content. For stronger identity protection, RID\cite{RID} moves this line toward practical deployment by replacing per-image optimization with a learned real-time generator. AdvCPG\cite{AdvCPG} combines identity-feature injection with gradient-based adversarial guidance during customized portrait generation, steering identity-related representations toward a target identity to mislead downstream face recognition models while preserving the desired visual appearance. StyleGuard\cite{StyleGuard} extends this representation-level view by explicitly attacking latent style features while modeling purification and upscaling attacks. Meanwhile, WaveGuard\cite{WaveGuard} broadens the same logic from source images to released generated outputs: although the perturbation is applied in output space, the effective target is the downstream latent or feature representation that a student model extracts from collected samples, making those outputs less useful for unauthorized distillation. Overall, this line of work develops from feature-space disruption for style mimicry to broader protection of identity, concepts, and student-side feature learning.
	
	\representationheading{Training dynamics and denoising representations}
	
	A second line of adversarial protection targets the internal learning signal used during unauthorized personalization. Instead of primarily targeting a semantic embedding, these methods aim to corrupt the denoising or optimization dynamics through which a downstream model learns from released samples. Anti-DreamBooth\cite{AntiDreamBooth} is the typical example: it protects portraits by optimizing perturbations against the DreamBooth denoising objective, thereby weakening downstream identity learning. MetaCloak\cite{MetaCloak} extends this direction through surrogate ensembles and bilevel meta-learning to improve transfer across models and input transformations. Score-Distillation Protection\cite{ScoreDistillationProtection} further argues that the encoder bottleneck provides a stronger and more stable protection surface than the denoiser, and optimizes a score-distillation objective accordingly. Meanwhile, MetaCloak-JPEG\cite{MetaCloakJPEG} adds differentiable JPEG and compression-aware optimization to model realistic acquisition pipelines. This line develops from directly attacking denoising-based personalization objectives to using surrogate-aware, bottleneck-aware, and compression-aware optimization for more transferable protection.
	
	\representationheading{Editing context, attention, and trajectory representations}
	
	A third line of work focuses on unauthorized image editing rather than downstream personalization. The task is to prevent an attacker from modifying a protected image through instruction-guided editing, inpainting, or context-based manipulation. Here, the targeted representations include editing latents, early denoising features, mask-conditioned trajectories, and multimodal attention pathways. EditShield\cite{EditShield} shifts the editing latent so that unauthorized edits produce mismatched or unrealistic results. DiffusionGuard\cite{DiffusionGuard} targets early denoising stages under unknown masks, while Anti-Inpainting\cite{AntiInpainting} attacks multi-level denoising features to improve robustness under unknown masks, seeds, and editing conditions. To defend against both personalization and editing misuse, Anti-Diffusion\cite{AntiDiffusion} expands the scope by combining prompt tuning with semantic disruption. DCT-Shield\cite{DCTShield} moves the carrier into JPEG-compatible frequency coefficients to improve survival under compression. While most methods focus on image-side perturbations, GuardDoor\cite{GuardDoor} departs from this setting by placing a protective mapping in a provider-controlled encoder, showing that system-level cooperation can supplement owner-only defenses. For more recent architectures, DeContext\cite{DeContext} adapts protection to modern DiT-based editors by weakening the multimodal attention pathways through which the source image conditions the edited output. Flux-Guard\cite{FluxGuard} provides an early flow-oriented example by coupling latent adversarial optimization with trajectory control for facial identity protection in FLUX-style editing.
	
	\representationheading{Evaluation, purification attacks, and model mismatch}
	
	For anti-training methods, evaluation should report degradation of downstream style imitation, identity personalization, concept learning, or student training, together with PSNR, SSIM, LPIPS, protection selectivity, runtime, and transfer across reuse methods. These targets can in principle appear in both diffusion and flow-based systems whenever they rely on similar encoders or latent interfaces. For anti-editing methods, evaluation should measure unauthorized edit success, preservation of original content and identity, robustness to unknown prompts and masks, resistance to compression and purification, and transfer across editing models. Diffusion-based methods often exploit discrete denoising states and early-stage UNet features, whereas FLUX-like systems expose different control surfaces through transformer attention and rectified-flow trajectories. 
	
	A common limitation is path dependence: protection may fail when the attacker bypasses the targeted pathway, reconstructs the input, applies adaptive preprocessing, or replaces the targeted representation altogether. ProtectionBenchmark\cite{ProtectionBenchmark} shows that adversarial protection should be assessed through a multi-objective trade-off among fidelity, protection strength, and robustness, rather than a single success metric. More importantly, recent attacks demonstrate that protected signals can often be weakened or removed before reuse. Off-the-Shelf Purification\cite{OffShelfPurification} shows that generic image-to-image generative models can already act as strong black-box purifiers, while Purify Once, Edit Freely\cite{PurifyOnce} highlights a more challenging model-mismatch setting in which the defender optimizes against one surrogate but the attacker purifies with another architecture before editing.

	\paragraph{Representation-level synthesis.}
	Across watermarking, backdoor attacks, and adversarial protection, the same representation can play very different functional roles. Despite these task differences, the practical effectiveness of representation-level methods is determined by several shared properties. A useful representation must be able to encode a task-specific signal, preserve that signal through generation or downstream optimization. These criteria help explain why some representation choices repeatedly appear across tasks. Current research is much more mature for diffusion models, whereas flow-based models remain relatively underexplored. A key next step is to develop methods specifically tailored to flow architectures rather than simply transferring diffusion-oriented designs.
	
	\subsection{Perception-Oriented Applications}\label{Perception-Oriented}
	\begin{table*}[t!]
		\centering
		\caption{Perception-Oriented Applications.}
		{\fontsize{6pt}{8pt}\selectfont
			\begin{tabular}{llcc}
				\hline\toprule
				\multicolumn{1}{c}{Downstream Tasks} & \multicolumn{1}{c}{Method} & \makecell[c]{Generative\\Paradigm} & Adaptation Strategy \\
				\midrule
				\multicolumn{4}{l}{\textit{Classification Tasks}}\\
				\midrule
				Image Classification & DDAE\cite{DDAE} & Diffusion & Leveraging intermediate representations\\
				Image Classification & GDC\cite{GDC} & Diffusion & Leveraging intermediate representations\\
				Image Classification & DifFormer\&DifFeed\cite{DifFormer} & Diffusion & Leveraging intermediate representations\\
				Image Classification & A. N. Juscafresa \textit{et al.}\cite{Diffusion-Plankton} & Diffusion & Leveraging intermediate representations\\
				Image Classification & RepFusion\cite{RepFusion} & Diffusion & Leveraging intermediate representations\\
				Image Classification & SymmFlow\cite{SymmFlow} & Flow Matching & Repurposing as direct predictors \\
				Image Classification & DFM\cite{DFM} & Flow Matching & Repurposing as direct predictors \\
				Image Classification & DiDiCM\cite{DiDiCM} & Diffusion & Repurposing as direct predictors \\
				Image Classification & ALIA\cite{ALIA} & Diffusion & Data synthesis \\
				Image Classification & Diff-Mix\cite{DiffMix} & Diffusion & Data synthesis \\
				Image Classification & AGA\cite{AGA} & Diffusion & Data synthesis \\
				Image Classification & DiffII\cite{DiffII} & Diffusion & Data synthesis \\
				Image Classification & Augmented Conditioning\cite{AugmentedConditioning} & Diffusion & Data synthesis \\
				Image Classification & SGD-Mix\cite{SGDMix} & Diffusion & Data synthesis \\
				Image Classification & OntoAug\cite{OntoAug} & Diffusion & Data synthesis \\
				Image Classification & GenMix\cite{GenMix} & Diffusion & Data synthesis \\
				Image Classification & Diffusion Curriculum\cite{DiffusionCurriculum} & Diffusion & Data synthesis \\
				Image Classification & TADA\cite{TADA} & Diffusion & Data synthesis \\
				Image Classification & StableRep\cite{StableRep} & Diffusion & Data synthesis \\
				Image Classification & SynCLR\cite{SynCLR} & Diffusion & Data synthesis \\
				Image Classification & Free-ATM\cite{Free-ATM} & Diffusion & Data synthesis \\
				Image Classification & DC\cite{DC} & Diffusion & Denoising-Evidence Classification \\
				Image Classification & MiPO\cite{MiPO} & Diffusion & Denoising-Evidence Classification \\
				Image Classification & MDC\cite{MDC} & Diffusion & Denoising-Evidence Classification \\
				Multi-Label Classification & Diff-Feat\cite{Diff-Feat} & Diffusion & Leveraging intermediate representations\\
				\midrule
				\multicolumn{4}{l}{\textit{Dense Visual Prediction Tasks}}\\
				\midrule
				Semantic Segmentation & DifFormer\&DifFeed\cite{DifFormer} & Diffusion & Leveraging intermediate representations\\
				Semantic Segmentation & RepFusion\cite{RepFusion} & Diffusion & Leveraging intermediate representations\\
				Semantic Segmentation & DDPM-Seg\cite{DDPM-Seg} & Diffusion & Leveraging intermediate representations\\
				Semantic Segmentation & DAAM\cite{DAAM} & Diffusion & Leveraging intermediate representations\\
				Semantic Segmentation & VPD\cite{VPD} & Diffusion & Leveraging intermediate representations\\
				Semantic Segmentation & MDM\cite{MDM} & Diffusion & Leveraging intermediate representations\\
				Semantic Segmentation & DDP\cite{DDP} & Diffusion & Repurposing as direct predictors \\
				Semantic Segmentation & DMP\cite{DMP} & Diffusion & Repurposing as direct predictors \\
				Semantic Segmentation & GenPercept\cite{GenPercept} & Diffusion & Repurposing as direct predictors \\
				Semantic Segmentation & SemFlow\cite{SemFlow} & Rectified Flow & Repurposing as direct predictors \\
				Semantic Segmentation & SymmFlow\cite{SymmFlow} & Rectified Flow & Repurposing as direct predictors \\
				Semantic Segmentation & Dataset Diffusion\cite{DatasetDiffusion} & Diffusion & Data synthesis\\
				Semantic Segmentation & FreeMask\cite{FreeMask} & Diffusion & Data synthesis\\
				Semantic Segmentation & ODISE\cite{ODISE} & Diffusion & Data synthesis\\
				Semantic Segmentation & Free-ATM\cite{Free-ATM} & Diffusion & Data synthesis \\
				Dichotomous Image Segmentation & GenPercept\cite{GenPercept} & Diffusion & Repurposing as direct predictors \\ 
				Dichotomous Image Segmentation & LawDIS\cite{LawDIS} & Diffusion & Repurposing as direct predictors \\
				Dichotomous Image Segmentation & FlowDIS\cite{FlowDIS} & Flow Matching & Repurposing as direct predictors \\
				In-context Segmentation & LDIS\cite{LDIS} & Flow Matching & Repurposing as direct predictors \\
				Depth Estimation & VPD\cite{VPD} & Diffusion & Leveraging intermediate representations\\
				Depth Estimation & ECoDepth\cite{ECoDepth} & Diffusion & Leveraging intermediate representations\\
				Depth Estimation & DDP\cite{DDP} & Diffusion & Repurposing as direct predictors \\
				Depth Estimation & Marigold\cite{Marigold} & Diffusion & Repurposing as direct predictors \\
				Depth Estimation & GenPercept\cite{GenPercept} & Diffusion & Repurposing as direct predictors \\
				Depth Estimation & DepthFM\cite{DepthFM} & Flow Matching & Repurposing as direct predictors \\
				Depth Estimation & CH3Depth\cite{CH3Depth} & Flow Matching & Repurposing as direct predictors \\
				Depth Estimation & Diffusion For Depth\cite{DiffusionForDepth} & Diffusion & Data synthesis\\
				\bottomrule\hline
			\end{tabular}
		}
		\label{Perception-Oriented Applications-1}
	\end{table*}
	\begin{table*}[t]
		\centering
		\caption{Perception-Oriented Applications (continued).}
		{\fontsize{6pt}{8pt}\selectfont
			\begin{tabular}{llcc}
				\hline\toprule
				\multicolumn{1}{c}{Downstream Tasks} & \multicolumn{1}{c}{Method} & \makecell[c]{Generative\\Paradigm} & Adaptation Strategy \\
				\midrule
				\multicolumn{4}{l}{\textit{Instance-Level Perception Tasks}}\\
				\midrule
				Panoptic Segmentation & ODISE\cite{ODISE} & Diffusion & Leveraging intermediate representations\\
				Referring Image Segmentation & VPD\cite{VPD} & Diffusion & Leveraging intermediate representations\\
				Referring Image Segmentation & LD-ZNet\cite{LD-ZNet} & Diffusion & Leveraging intermediate representations\\
				Referring Image Segmentation & RLFSeg\cite{RLFSeg} & Rectified Flow & Repurposing as direct predictors \\
				Object Detection & DiffusionDet\cite{DiffusionDet} & Diffusion & Repurposing as direct predictors \\
				Object Detection & DifFormer\&DifFeed\cite{DifFormer} & Diffusion & Leveraging intermediate representations\\
				Object Detection & DiffusionInst\cite{DiffusionInst} & Diffusion & Repurposing as direct predictors \\
				Object Detection & Free-ATM\cite{Free-ATM} & Diffusion & Data synthesis \\
				Object Detection & DALL-E Detection\cite{DALL-E-Detection} & Diffusion & Data synthesis \\
				Object Detection & MosaicFusion\cite{MosaicFusion} & Diffusion & Data synthesis \\
				Object Detection & DiffusionEngine\cite{DiffusionEngine} & Diffusion & Data synthesis \\
				Object Detection & ReCon\cite{ReCon} & Diffusion & Data synthesis \\
				Keypoint Detection & RepFusion\cite{RepFusion} & Diffusion & Data synthesis \\
				Instance Segmentation & OC-DiT\cite{OC-DiT} & Diffusion & Repurposing as direct predictors \\
				Instance Segmentation & MosaicFusion\cite{MosaicFusion} & Diffusion & Repurposing as direct predictors \\
				Open-Vocabulary Object Segmentation & Grounded Diffusion\cite{GroundedDiffusion} & Diffusion & Data synthesis\\
				\midrule
				\multicolumn{4}{l}{\textit{Annotation-Scarce Tasks}}\\
				\midrule
				Few-shot Image Classification & DAFusion\cite{DAFusion} & Diffusion & Data synthesis\\
				Few-shot Image Classification & DIAGen\cite{DIAGen} & Diffusion & Data synthesis\\
				Label-efficient Semantic Segmentation & DreamTeacher\cite{DreamTeacher} & Diffusion & Leveraging intermediate representations\\
				Incremental Few-Shot Semantic Segmentation & iFSS-Diff\cite{iFSS-Diff} & Diffusion & Repurposing as direct predictors \\
				Weakly-supervised Semantic Segmentation & ScribbleGen\cite{ScribbleGen} & Diffusion & Data synthesis \\
				Open-vocabulary Semantic Segmentation & DreamMask\cite{DreamMask} & Diffusion & Data synthesis\\
				Few-shot Instance Segmentation & MaskDiff\cite{MaskDiff} & Diffusion & Repurposing as direct predictors \\
				Long-tailed Instance Segmentation & TMI\cite{TMI} & Diffusion & Data synthesis\\
				Long-tailed Instance Segmentation & Gen-n-Val\cite{Gen-n-Val} & Diffusion & Data synthesis\\
				Open-vocabulary Panoptic Segmentation & DreamMask\cite{DreamMask} & Diffusion & Data synthesis\\
				Long-tailed Object Detection & DiverGen\cite{DiverGen} & Diffusion & Data synthesis\\
				Long-tailed Object Detection & InstaDA\cite{InstaDA} & Diffusion & Data synthesis\\
				Long-tailed Object Detection & Gen-n-Val\cite{Gen-n-Val} & Diffusion & Data synthesis\\
				
				\bottomrule\hline
			\end{tabular}
		}
		\label{Perception-Oriented Applications-2}
	\end{table*}
	
	The representation capabilities of diffusion models and flow-based models are not confined to the generative task itself. Through training aimed at generating high-quality images, these models learn rich visual representations and gradually develop a structured understanding of visual scenes. This chapter focuses on the application of diffusion models and flow-based models to three representative perception tasks: image classification (Sec. \ref{Image Classification Task}), dense visual prediction (Sec. \ref{Dense Visual Prediction Task}, including semantic segmentation, dichotomous segmentation, and depth estimation), and instance-level perception (Sec. \ref{Instance-Level Perception Task}, including instance segmentation, panoptic segmentation, and object detection). We additionally summarize works dedicated to annotation-scarce scenarios (Sec. \ref{Annotation-Scarce Task}). Relevant approaches can be broadly categorized into three distinct directions: (1) extracting relevant representations from intermediate layers of pre-trained models and using them as a foundation for training downstream perception models; (2) directly adapting diffusion or flow-based models to produce perception task outputs, such as segmentation masks or depth maps; and (3) leveraging generative models to synthesize training data to improve the performance of downstream perception models, particularly when dealing with long-tailed or class-imbalanced datasets. The methods introduced in this section are summarized in Table \ref{Perception-Oriented Applications-1} and \ref{Perception-Oriented Applications-2}.
	\subsubsection{Image Classification Task}\label{Image Classification Task}
	Classification is one of the most direct ways to examine whether a generative model learns discriminative visual representations. In this setting, generation is not only used to produce realistic images, but also to improve the class coverage, robustness, or task interface of a classifier. Most methods fall into the three mainstream adaptation strategies summarized at the beginning of this chapter. Beyond these, there is an independent technical route, namely Denoising-Evidence Classification, which treats the conditional diffusion model as an implicit discriminator by comparing the denoising errors under different labels. In the following, we introduce relevant methods according to different paradigms.
	\paragraph{Leveraging the Intermediate Representations of the Model}
	Some methods attempt to leverage pre-trained diffusion models as frozen feature extractors, directly utilizing their internal intermediate activations for discriminative tasks. These methods do not require generating additional samples or modifying model architectures. Instead, they directly mine the structured representations that diffusion models spontaneously learn during generative training.
	
	DDAE\cite{DDAE} is the foundational work in this direction. Its core idea is to use a pre-trained diffusion model (DDPM) as a feature extractor. An input image is first noised at a specific timestep $t$ and then fed into the U-Net. Activations are extracted from its intermediate layers and pooled globally to obtain fixed-dimensional feature vectors. A linear classifier is then trained on the vectors for evaluation. This work systematically demonstrates for the first time that generative pre-training alone, without contrastive learning or additional discriminative objectives, enables diffusion models to learn strongly linearly separable visual representations. 
	
	GDC\cite{GDC} builds upon this foundation by introducing an attention-based classification head, which significantly improves performance. Through CKA analysis, this work further reveals a high similarity between the deep representations of diffusion models and the discriminative representations of ResNet and ViT, providing a representational explanation for why diffusion models possess discriminative capabilities.
	
	DifFormer and DiffFeed\cite{DifFormer} represent a further extension of GDC. This work observes that discriminative information in diffusion models is not concentrated in a single block or timestep, but rather distributed across different locations. To address this, they propose DiffFormer, which employs a Transformer to fuse features from multiple timesteps and multiple U-Net blocks, achieving cross-timestep and cross-block information aggregation. They also propose DiffFeed, which leverages the symmetric encoder-decoder structure of U-Net by feeding decoder features from the first forward pass back into the encoder during a second forward pass, achieving feature enhancement at the cost of only two forward passes. Meanwhile, this work extends the application scope to other downstream tasks including semantic segmentation and object detection, validating the potential of diffusion models as general-purpose representation learners.
	
	Diff-Feat\cite{Diff-Feat} and A. N. Juscafresa \textit{et al.}\cite{Diffusion-Plankton} further extend this paradigm to cross-modal multi-label classification and fine-grained plankton classification, respectively. Both works point out that the quality of internal representations strongly depends on the selected network layer and noise level. Among them, Diff-Feat reveals through extensive experiments that the twelfth layer of DiT is the optimal extraction position for representations.
	
	In contrast to the aforementioned methods that directly extract frozen features, RepFusion\cite{RepFusion} transfers the intermediate representations of diffusion models to lightweight student networks via knowledge distillation. RepFusion establishes a theoretical connection between diffusion models and denoising autoencoders, demonstrating a representation-regularization trade-off across different timesteps. To address the challenge of optimal timestep selection, they introduce a reinforcement learning strategy to dynamically select the most suitable distillation timestep for each sample. This method can also be applied to other tasks such as semantic segmentation and keypoint detection.
	
	\paragraph{Repurposing Generative Models as Direct Predictors}
	Some works apply flow-based methods or diffusion models directly to obtain task outputs instead of generating complete images. The input image is first mapped to a visual representation. A conditional generative process then transforms noise into the prediction of classification.
	
	DFM~\cite{DFM} formulates image classification and object detection as conditional transport. Given an image representation $h(x)$, it learns a time-dependent vector field that transports an initial noise sample toward a task target,
	
	\begin{equation}
		z_0 \sim p_0,
		\qquad
		\frac{\mathrm{d}z_t}{\mathrm{d}t}
		=
		v_{\theta}\left(z_t,t\mid h(x)\right),
		\qquad
		z_1 \approx \tau_y,
	\end{equation}
	
	where $\tau_y$ represents a class embedding for classification or a geometric target for object detection.
	
	DFM attaches several local flow predictors to different blocks of a shared CNN or Vision Transformer. Each predictor is trained with a local Flow Matching objective, and their outputs are combined to produce the final prediction. The local structure allows sequential updates to reduce activation memory or parallel updates to improve computational efficiency.
	
	DiDiCM~\cite{DiDiCM} uses a discrete diffusion process instead of a continuous flow. It models the conditional distribution of class labels given an input image. The diffusion process can operate on class-probability vectors or directly on discrete labels.
	
	DiDiCM begins from a noisy or uncertain label state and gradually refines it through several reverse steps. This differs from a standard classifier, which produces a prediction through a single fixed mapping. The number of reverse steps provides a trade-off between computation and prediction quality. The method is designed for settings with corrupted inputs, limited training data, or high prediction uncertainty.
	
	\paragraph{Synthetic Data for Classification}
	Much of the recent research has focused on using synthetic data for classification. The motivation is that conventional transformations such as cropping, rotation, or color jitter only cover low-level variation, while many classification errors come from missing object poses, backgrounds, styles, or tail categories. According to how synthetic data is used by the classifier, existing works can be divided into \textbf{sample-level augmentation}, \textbf{distribution-controlled augmentation}, and \textbf{synthetic-data-driven representation learning}.
	
	\textbf{Sample-level augmentation} directly generates additional images or features for classifier training.
	Early generative augmentation mainly addresses data scarcity by learning class-preserving variations from limited examples.
	GAN-based augmentation methods such as DAGAN~\cite{DAGAN} and BAGAN~\cite{BAGAN} synthesize class-preserving or minority-class samples for few-shot and long-tailed recognition.
	Delta-encoder~\cite{DeltaEncoder} performs a related idea in feature space by learning intra-class deformation vectors.
	With pretrained diffusion models, augmentation shifts from training a small generator on the target dataset to prompting or conditioning a strong image prior.
	ALIA~\cite{ALIA} automatically describes training domains and edits contextual factors while filtering label-corrupting results.
	Diff-Mix~\cite{DiffMix} performs inter-class diffusion mixup and uses mixed supervision to balance foreground faithfulness and contextual diversity.
	AGA~\cite{AGA} combines language, segmentation, and diffusion modules to preserve foreground subjects and diversify backgrounds.
	The main issue is no longer only whether images look realistic, but whether the synthetic variation matches the classifier's missing modes without introducing label noise.
	
	\textbf{Distribution-controlled augmentation} focuses on generating the right variation rather than simply increasing the number of samples.
	The central motivation is that uncontrolled synthesis may change class identity, amplify spurious correlations, or add easy samples that contribute little to decision boundaries.
	DiffII~\cite{DiffII} interpolates same-class diffusion inversions to keep generated samples inside the target class manifold, while Augmented Conditioning~\cite{AugmentedConditioning} uses conventional augmentations as diffusion conditions to turn simple perturbations into richer in-domain images.
	SGD-Mix~\cite{SGDMix} and OntoAug~\cite{OntoAug} explicitly preserve foreground semantics while diversifying backgrounds through saliency, mask, or ontology constraints.
	GenMix~\cite{GenMix} instead combines prompt-guided generative editing with image and fractal mixing to improve in-domain and cross-domain classification.
	Classifier-aware strategies further decide which synthetic samples should be generated or used according to curriculum design or early training dynamics.
	Diffusion Curriculum~\cite{DiffusionCurriculum} organizes the synthetic-to-real training order, whereas TADA~\cite{TADA} selects slow-learnable examples for targeted diffusion augmentation.
	These methods share the view that useful generation should expand the decision-relevant part of the distribution while controlling semantic drift and sample redundancy.
	
	\textbf{Synthetic-data-driven representation learning} treats generated images as a broad pretraining distribution rather than a local supplement to a target dataset.
	In this setting, the generator provides large-scale visual diversity before a downstream classifier is trained, so the key question becomes whether synthetic images can support transferable representation learning.
	StableRep~\cite{StableRep} uses multiple images from the same text prompt as positive pairs, turning prompt-level semantic consistency into a contrastive learning signal.
	SynCLR~\cite{SynCLR} scales the pipeline with language-generated captions and large synthetic image collections, showing that representations learned from generated data can transfer to real classification tasks.
	Free-ATM\cite{Free-ATM} leverages synthetic images generated by diffusion models, and simultaneously uses the cross-attention maps produced during generation as free pixel-level supervision signals to improve representation learning. The method can also be broadly applied to downstream vision tasks such as object detection and semantic segmentation.
	
	Compared with sample-level augmentation, synthetic-data-driven representation learning moves the bottleneck from per-class sample generation to prompt design, filtering, and distribution coverage.
	
	\paragraph{Denoising-Evidence Classification with Diffusion Models} 
	Unlike the three mainstream adaptation strategies summarized at the beginning of this section, some works directly treat a conditional diffusion model itself as an implicit discriminator, performing classification by comparing the conditional likelihoods under different labels. Specifically, they compare how well the conditional diffusion model explains the input image under each candidate label. The main assumption is that the model should predict the injected noise more accurately when conditioned on the correct label.
	
	For each candidate label, the input image is corrupted at several diffusion timesteps and with several noise samples. The diffusion model then predicts the injected noise while being conditioned on the corresponding class description or class embedding. The prediction errors are averaged across the sampled timesteps and noise instances. Different timesteps may also be assigned different weights. The class with the lowest average denoising error is selected as the final prediction.
	
	DC~\cite{DC} applies this idea to class-conditional and text-to-image diffusion models. For each candidate label, it evaluates the denoising error over multiple timesteps and noise samples. The condition that gives the lowest average prediction error is selected as the predicted class.
	
	DC does not require an additional classification head or task-specific generative training. It can also perform zero-shot classification by converting class names into text conditions. However, its inference cost can be high because the diffusion model must be evaluated several times for every candidate class.
	
	MiPO~\cite{MiPO} studies the tendency of diffusion classifiers to perform better on concepts that are common in the pretrained generative distribution. It assigns larger rewards to generated samples that better cover rare or low-density concepts and then adapts the diffusion model using these preference signals.
	
	MiPO uses parameter-efficient adaptation and preference optimization to improve minority coverage without requiring additional downstream images or an external reward model. Better coverage of minority concepts in the generative distribution also improves their zero-shot classification performance.
	
	Building upon DC, MDC\cite{MDC} further observes that denoising errors are not uniformly distributed across the image, and errors concentrated in the foreground region carry more discriminative meaning than those in the background. Based on this insight, MDC leverages both cross-attention and self-attention maps from the text-to-image diffusion model as a soft mask to weigh the denoising errors, forcing the model to focus on the object of interest. Additionally, MDC introduces a structural distance that compares the structural information extracted from the original image and the attention map, capturing differences in the shape and size of objects. By integrating semantic and structural distances, MDC assigns the image to the category with the minimum combined distance.
	
	\subsubsection{Dense Visual Prediction Task}\label{Dense Visual Prediction Task}
	Different from image classification, which models a global holistic representation of the entire image, dense visual prediction tasks require models to assign semantic or geometric labels to every pixel of the input image, encompassing representative tasks such as semantic segmentation, dichotomous segmentation, and monocular depth estimation. These tasks require the representations to preserve high-resolution spatial details for localization accuracy and encode sufficient global context for semantic or geometric reasoning as well.
	\paragraph{Leveraging the Intermediate Representations of the Model} The U-Net architecture of diffusion models, with its hierarchical encoder-decoder design and skip connections, naturally provides a spectrum of feature maps spanning from fine-grained texture to semantic abstraction, making it inherently suitable for dense prediction tasks. Therefore, similar to DDAE\cite{DDAE}, some works have attempted to leverage pre-trained diffusion models by extracting intermediate features and applying them to dense visual prediction tasks. DDPM-Seg\cite{DDPM-Seg}, as one of the earliest efforts, adopts a strategy closely analogous to DDAE. It extracts activation maps from multiple intermediate layers of the U-Net decoder of a pre-trained DDPM, upsamples and concatenates them, and then trains a lightweight classifier on top to perform semantic segmentation.
	
	Another line of works focus on the role of cross-attention maps within diffusion models. DAAM\cite{DAAM} systematically reveals that the cross-attention maps within Stable Diffusion\cite{StableDiffusion} naturally encode class-to-position bindings. By aggregating cross-attention maps from various layers of the model, DAAM achieves semantic segmentation without any training. VPD\cite{VPD} further utilizes the cross-attention maps by concatenating them with multi-scale feature maps from the U-Net decoder at the same resolution along the channel dimension, using them as explicit semantic guidance fed into the prediction head. Through lightweight U-Net fine-tuning and prediction head training, VPD achieves competitive performance across multiple downstream tasks, including semantic segmentation, depth estimation, and referring image segmentation. Similar to VPD, ECoDepth\cite{ECoDepth} also injects task-relevant conditions into the diffusion model. It points out that text descriptions tend to focus on large salient objects; to enable the model to attend to small objects and details in depth estimation, ECoDepth employs a frozen pre-trained ViT to extract class probability vectors, projects them through an MLP, and injects them into the cross-attention layers of the Stable Diffusion U-Net. This design enables the model to outperform VPD on depth estimation and generalize effectively to other datasets.
	
	Different from the aforementioned methods that directly leverage features from pre-trained models, MDM\cite{MDM} redesigns and trains a diffusion encoder from scratch with dense prediction tasks as the target. It points out that there is an inherent misalignment between the denoising generation objective of traditional diffusion models and downstream segmentation tasks: generative models need to capture high-frequency texture details to synthesize realistic images, while dense prediction tasks such as segmentation rely more on low-frequency structural information. Based on this insight, MDM replaces the Gaussian noise addition in traditional diffusion models with a dynamic masking mechanism. Meanwhile, MDM introduces the Structural Similarity Index (SSIM) loss to replace the Mean Squared Error (MSE) loss, forcing the encoder to prioritize preserving geometric structure and semantic boundaries during reconstruction. After pre-training, the U-Net encoder is frozen and serves as a feature extraction backbone, attaching a lightweight MLP prediction head on top of its multi-scale features to perform segmentation tasks. 
	
	The core idea of the above methods all lies in extracting learned representations from diffusion models to serve downstream tasks, with the key distinction being that DDPM-Seg, DAAM, VPD, and ECoDepth leverage internal features from off-the-shelf pre-trained generative models, while MDM redesigns the pre-training objective specifically for perception tasks before extraction. More importantly, MDM demonstrates that the representation learning capability of diffusion models can exist independently of their generative capacity. By redesigning the pre-training objective, a diffusion architecture that completely lacks image generation capability can still learn high-quality representations for dense prediction. If we further treat the entire denoising process of the diffusion model itself as the inference pathway and directly replace its output target, this constitutes another more thorough technical route: repurposing generative models as dense predictors directly.
	\paragraph{Repurposing Generative Models as Direct Predictors}
	Multiple recent works have focused on directly modifying the output target of diffusion models, such that they no longer generate RGB images but instead directly regress dense prediction labels, including depth maps and semantic masks, from the input image. DDP\cite{DDP} is one of the early representatives of this direction, replacing the standard noise prediction loss of conditional diffusion models with task-specific supervision signals (e.g., cross-entropy loss for semantic segmentation and L1 loss for depth estimation). Architecturally, DDP decouples the image encoder from a lightweight map decoder, allowing the image encoder to run only once while the diffusion iterations are performed repeatedly on the decoder, significantly improving efficiency.
	
	Similarly, Marigold\cite{Marigold} repurposes Stable Diffusion, a large-scale pre-trained text-to-image diffusion model, for monocular depth estimation. Its core design principle is to keep the latent space of the pre-trained model completely intact, modifying only the input layer of the U-Net to concatenate the image and depth latents, and fine-tuning the entire U-Net. Marigold is trained using only synthetic data, yet achieves state-of-the-art zero-shot generalization across multiple real-world datasets, demonstrating the transferability of large-scale generative priors to geometric understanding tasks. DMP\cite{DMP} takes a different perspective, addressing the misalignment between the stochasticity of diffusion models and the deterministic nature of dense prediction tasks by designing a deterministic diffusion process. To be specific, DMP formulates the diffusion path as a chain of interpolations between the input image and the output label, such that the reverse process becomes a deterministic de-interpolation from the input image to the output label. Combined with LoRA fine-tuning, DMP achieves generalizable dense prediction.
	
	Building upon previous work, GenPercept\cite{GenPercept} further conducts systematic ablation studies to uncover the key design factors in the repurposing process: (1) by setting the hyperparameters of the diffusion scheduler to specific values, multi-step denoising can be simplified to single-step deterministic inference without performance loss; (2) the U-Net is the primary carrier of perceptual priors, and freezing the U-Net or using only LoRA fine-tuning underperforms full fine-tuning significantly; (3) timesteps and text prompts contribute negligibly to deterministic perception tasks. Based on these findings, GenPercept extends the repurposing framework to five dense perception tasks, including depth estimation, surface normal estimation, semantic segmentation, dichotomous segmentation, and image matting.
	
	Other works further explore the applicability of diffusion models to other downstream tasks. LDIS\cite{LDIS} extends the repurposing idea to in-context segmentation. Given a reference image-mask pair as visual prompts, LDIS leverages the LDM to generate segmentation masks for the query image in one or a few sampling steps, without requiring text instructions or additional refinement networks, validating the potential of diffusion models in cross-task generalization scenarios.  LawDIS\cite{LawDIS} recasts dichotomous image segmentation(DIS) as an image-conditioned mask generation task within a latent diffusion model, enabling language-guided initial mask generation and user-controlled window-based refinement for high-precision foreground segmentation.
	
	Although the diffusion-based repurposing works discussed above differ in task settings and implementation details, they share a common characteristic: they retain the denoising architecture of diffusion models. In contrast, recent work has pointed out that the stochastic denoising process of diffusion models is inherently misaligned with the deterministic mapping required by dense prediction tasks, and that the deterministic ordinary differential equation (ODE) paradigm of flow matching offers greater advantages in such scenarios\cite{SemFlow, FlowDIS, RLFSeg}. We next introduce flow-based methods for dense visual prediction tasks.
	
	SemFlow\cite{SemFlow} is an early exploration in this direction for semantic segmentation. Built upon Rectified Flow\cite{RF}, it treats semantic segmentation and semantic image synthesis as a pair of inverse problems and establishes a unified bidirectional mapping framework, enabling reversible translation between the image domain and the semantic domain with a single set of shared parameters. Building upon this work, SymmFlow\cite{SymmFlow} further adopts a symmetric flow matching training objective that independently perturbs images and semantic labels toward their respective noise distributions. By introducing a label dequantization strategy that relaxes the dimensional constraint between images and masks, SymmFlow extends the unified framework from segmentation and generation to image classification, which has been mentioned in Sec. \ref{Image Classification Task}.
	
	In the domain of depth estimation, DepthFM\cite{DepthFM} makes the first attempt to apply flow matching to monocular depth estimation, formulating the task as a direct transport between the image distribution and the depth distribution, rather than iterative denoising from Gaussian noise. With the initialization from a pre-trained diffusion model, DepthFM achieves single-step or two-step inference, significantly reducing inference latency while maintaining competitive accuracy. CH3Depth\cite{CH3Depth} further refines the objective function of flow matching by reformulating it as direct iterative inversion (InDI), combined with a non-uniform sampling strategy, achieving a better balance between accuracy and efficiency. Its proposed Latent Temporal Stabilizer (LTS) further extends the model to video depth estimation, delivering leading zero-shot performance across multiple image and video benchmarks.
	
	Furthermore, flow matching has also been applied to the Dichotomous Image Segmentation (DIS) task, which demands high boundary precision. FlowDIS\cite{FlowDIS} learns a time-dependent vector field that transports the image distribution directly to the corresponding mask distribution, and optionally conditioned on text prompts for language-guided segmentation. Its proposed position-aware instance pairing training strategy effectively enhances language controllability in multi-object scenes, significantly surpassing prior SOTA methods on DIS benchmarks.
	
	\paragraph{Synthetic Data for Dense Prediction}
	Some works directly leverage the generative capacity of diffusion models to produce annotated training data, thereby augmenting existing datasets and enhancing the generalization and robustness of downstream perception models. The core rationale of such approaches lies in the rich visual priors acquired by diffusion models during large-scale pre-training: through proper conditioning designs, these models can generate diverse data to provide complementary training signals for discriminative models.
	
	In semantic segmentation, DatasetDiffusion\cite{DatasetDiffusion} synthesizes large-scale pixel-level semantic segmentation datasets via class prompts and cross-attention exponentiation strategies, demonstrating the feasibility of synthetic data in segmentation tasks. FreeMask\cite{FreeMask} further refines this paradigm by generating images conditioned on real semantic masks, re-sampling based on mask difficulty to focus on more challenging samples, while systematically filtering out artifacts in the generated images, thereby improving both the quality and training efficiency of synthetic data.
	
	As for depth estimation, DiffusionForDepth\cite{DiffusionForDepth} employs a text-to-image conditional diffusion model to transform simple scene images into challenging scenarios involving adverse lighting, rain/snow, or non-Lambertian surfaces, while preserving the original 3D structure to ensure label correctness. Through self-distillation fine-tuning, this method significantly enhances the robustness of pre-trained depth models under challenging conditions, demonstrating the potential of synthetic data in geometric perception tasks.
	
	The above works demonstrate that densely annotated data synthesized by diffusion models hold significant value in expanding training distributions and alleviating annotation bottlenecks.
	\subsubsection{Instance-Level Perception Task}\label{Instance-Level Perception Task}
	Unlike dense prediction tasks that assign semantic labels to each pixel, instance-level perception tasks require models to not only recognize object categories but also distinguish individual instances within the same category, outputting bounding boxes or instance-level masks. Representative tasks include object detection, instance segmentation, panoptic segmentation, and referring image segmentation. These tasks impose higher demands on both the instance discrimination capability and spatial localization accuracy of visual representations, and consequently, the utilization of internal representations from diffusion models differs from that in dense prediction settings. In this section, we follow the three similar technical paradigms as in the previous section to systematically review the application of diffusion models and flow-based models to instance-level perception tasks.
	\paragraph{Leveraging the Intermediate Representations of the Model}
	In line with the spirit of DDAE and DAAM, some researchers have also attempted to extract internal representations from frozen pre-trained diffusion models to serve open-vocabulary and text-driven instance segmentation. ODISE\cite{ODISE} reveals through k-means clustering that the internal representations of text-to-image diffusion models exhibit high semantic discriminability and spatial localization accuracy. Accordingly, ODISE extracts internal features from the frozen U-Net of Stable Diffusion via a single forward pass, substitutes the real text condition with an implicit text embedding to accommodate unannotated inference scenarios, and feeds the extracted features into a Mask2Former decoder to generate masks, effectively achieving open-vocabulary panoptic segmentation and generalizing across tasks to open-vocabulary object detection and open-world instance segmentation.
	
	LD-ZNet\cite{LD-ZNet} focuses on referring expression segmentation, leveraging LDM latent features and VQGAN features fused via cross-attention at specific diffusion timesteps (300?500) to achieve text-driven instance-level segmentation, further demonstrating the potential of diffusion model internal representations in cross-modal instance perception. The above methods accomplish instance-level perception tasks without retraining the encoder by freezing pre-trained diffusion models and extracting their internal features, demonstrating the generality of diffusion model representations. However, these approaches are inherently constrained by the capability of the pre-trained models themselves; their performance heavily depends on the encoding quality of instance boundaries within the diffusion model's internal representations, and they still face challenges when dealing with complex scenes or heavy object overlap.
	\paragraph{Repurposing Generative Models as Direct Predictors}
	Another line of work aims to replace the output target of the model with bounding boxes or instance masks, enabling diffusion or flow-based models to directly regress instance-level predictions from the input image. DiffusionDet\cite{DiffusionDet} is an early exploration of this idea in object detection, formulating the generation of detection boxes as a denoising process from noisy boxes to ground-truth boxe. It is trained with ground-truth boxes as the learning target and infers by progressively denoising from random noise boxes to directly predict bounding boxes. Building upon this, DiffusionInst\cite{DiffusionInst} extends the same repurposing paradigm to instance segmentation by generating instance mask vectors. In the zero-shot setting, OC-DiT\cite{OC-DiT} leverages a conditional latent diffusion model, combining DINOv2 features with cross-attention from CAD-rendered multi-view images, to generate initial instance masks and achieve zero-shot generalization through independent refinement.
	
	Beyond the approaches discussed above, RLFSeg\cite{RLFSeg} points out that one-to-one deterministic mapping is more suitable for tasks with deterministic output requirements. It adopts Rectified Flow\cite{RF} instead of diffusion models for referring expression segmentation, and refines masks with SAM\cite{SAM} to optimize prediction quality. Furthermore, RLFSeg can also be generalized to semantic segmentation tasks in a zero-shot manner. The above works demonstrate that the paradigm of repurposing generative models as direct predictors can be naturally extended from dense prediction tasks to instance-level tasks.
	\paragraph{Synthetic Data for Instance-Level Perception}
	The annotation cost of instance-level tasks is substantially higher than that of semantic segmentation, as labeling bounding boxes or pixel-wise masks for object detection and instance segmentation requires extensive manual effort. Therefore, leveraging diffusion models to synthesize data with instance-level annotations has become a highly attractive technical direction. DALL-E Detection\cite{DALL-E-Detection} is one of the early explorations in this direction, adopting a step-by-step composition strategy that first generates foreground objects and their corresponding masks, then synthesizes backgrounds based on textual descriptions, and finally composes them into complete detection data. GroundedDiffusion\cite{GroundedDiffusion} combines Stable Diffusion with a grounding module to synthesize open-vocabulary segmentation data. Building upon this, DiffusionEngine\cite{DiffusionEngine} designs an additional adapter that trains a detection head from U-Net features based on the DINO framework, enabling scalable detection data generation. MosaicFusion\cite{MosaicFusion} leverages Stable Diffusion to generate images containing multiple objects and aggregates the internal cross-attention maps to produce corresponding instance masks, synthesizing large-scale instance segmentation data without manual annotation. ReCon\cite{ReCon} introduces a region-aligned cross-attention mechanism to prevent semantic leakage, replacing the post-generation filtering of low-quality data adopted by other methods with in-process rectification during sampling, thereby improving both the quality and the generation speed of synthetic data.
	\subsubsection{Annotation-Scarce Task}\label{Annotation-Scarce Task}
	Most methods discussed in the preceding sections assume the availability of annotated data for training. In practice, however, acquiring high-quality annotations is often prohibitively expensive, especially for tasks such as semantic segmentation, instance segmentation, and object detection. When annotated data is extremely scarce, the generative capabilities of diffusion models offer an attractive alternative: synthesizing annotated training data to augment the existing data distribution, thereby improving model performance in few-shot, long-tailed, weakly-supervised, and open-vocabulary scenarios. In this section, we focus on works that are specifically designed for annotation-scarce scenarios or make relevant theoretical contributions. Note that although the previous sections also include works on synthetic data for training, their primary motivation is not to address data scarcity, and thus they are not included in this section. Interestingly, the works in this section can still be organized according to the same three paradigms introduced earlier.
	
	\paragraph{Leveraging the Intermediate Representations of the Model}
	In annotation-scarce scenarios, some researchers have attempted to extract semantic representations from pre-trained diffusion models to alleviate the difficulties caused by limited annotations. DreamTeacher \cite{DreamTeacher} uses the intermediate features of a pre-trained diffusion model as teacher signals, distilling them into lightweight image backbones such as ResNet\cite{ResNet} via MSE loss and attention transfer loss. This framework can be applied to semi-supervised learning, classification, instance segmentation, and label-efficient semantic segmentation, enabling small networks to inherit the rich visual priors learned by large-scale diffusion models, thus maintaining strong performance even when annotations are scarce.
	
	\paragraph{Repurposing Generative Models as Direct Predictors}
	Other works replace the output target of diffusion or flow-based models with prediction labels and introduce specialized adaptations, enabling generative inference to perform predictions in annotation-scarce scenarios. MaskDiff \cite{MaskDiff} employs diffusion models to directly generate instance masks, demonstrating in few-shot instance segmentation that generative methods exhibit stronger robustness than traditional discriminative approaches when data is extremely scarce. iFSS-Diff \cite{iFSS-Diff} learns a dedicated embedding for each base class and generates masks, achieving incremental few-shot semantic segmentation that allows the model to segment new classes incrementally after training on base classes.
	
	\paragraph{Synthetic Data for Representation Learning}
	Different from the previous two paradigms, more works take a more fundamental approach from the data level, leveraging the conditional generation capability of diffusion models to synthesize annotated training data for augmenting existing datasets. This has become the most active technical direction in annotation-scarce scenarios.
	
	ScribbleGen \cite{ScribbleGen}, as an early work in this direction, preliminarily explores the potential of diffusion models in weakly-supervised scenarios. It generates synthetic data based on diffusion models, effectively improving semantic segmentation performance under sparse scribble annotations.For few-shot image classification, DA-Fusion~\cite{DAFusion} uses pretrained diffusion models for semantic image-to-image augmentation in few-shot settings, while DIAGen~\cite{DIAGen} emphasizes semantically diverse augmentation for few-shot learning. In open-vocabulary scenarios, DreamMask \cite{DreamMask} leverages GPT-4o to generate new category names, refines descriptions via LLM, synthesizes images with diffusion models, and applies CLIP \cite{CLIP} similarity filtering and SAM \cite{SAM} uncertainty filtering to achieve high-quality open-vocabulary panoptic segmentation data synthesis.
	
	In addition, many works focus on addressing long-tailed distribution problems. DiverGen \cite{DiverGen} adopts a multi-level diversity enhancement strategy, also utilizing SAM and CLIP for post-processing of generated data, effectively improving instance segmentation performance on long-tailed categories. TMI \cite{TMI} employs a complementary synthesis strategy combining text-to-image and image-to-image generation, using the T2I branch to provide scene diversity and the I2I branch to insert high-confidence instances into real scenes to maintain semantic coherence, effectively alleviating data scarcity in long-tailed scenarios.
	
	Recent works have gradually shifted toward reducing the failure rate of generated data. InstaDA \cite{InstaDA} proposes a dual-agent system that introduces LLM and Prompt Rethink mechanisms to enhance data diversity, and generates new instances based on training images to address the underutilization of annotated data. Additionally, InstaDA introduces a self-correction mechanism to cope with the issue of high image discard rates. Gen-n-Val \cite{Gen-n-Val} leverages Layer Diffusion to generate foreground and alpha channels, and employs VLLM for data validation and filtering, significantly reducing the failure rate of synthetic data, with applications in long-tailed instance segmentation and open-vocabulary detection.
	
	\subsection{Generalist and Unified Applications}\label{Generalist and Unified}
	\subsubsection{Clustering Analysis}\label{Clustering Analysis}
	\begin{table*}[t]
		\centering
		\caption{Clustering-related applications of diffusion and flow-based models.}
		\label{tab:clustering-applications}
		{\fontsize{7pt}{10pt}\selectfont
			\begin{tabular}{llcl}
				\hline\toprule
				\multicolumn{1}{c}{\textbf{Clustering Direction}}
				& \multicolumn{1}{c}{\textbf{Method}}
				& \multicolumn{1}{c}{\makecell{\textbf{Generative}\\\textbf{Paradigm}}}
				& \multicolumn{1}{c}{\textbf{Adaptation Strategy}} \\
				\midrule
				
				\multicolumn{1}{l}{\makecell[l]{Generative Clustering with\\Diffusion Models}}
				& ClusterDDPM\cite{ClusterDDPM}
				& Diffusion
				& Joint diffusion learning and EM clustering \\
				\midrule
				
				\multirow{4}{*}{\makecell[l]{Diffusion-Based\\Assignment and\\Representation Clustering}}
				& CLUDI\cite{CLUDI}
				& Diffusion
				& Diffusion-based cluster assignment \\
				
				& DiFiC\cite{DiFiC}
				& Diffusion
				& Semantic-condition clustering \\
				
				& DiEC~\cite{diec}
				& Diffusion
				& Intermediate diffusion-feature clustering \\
				
				& Subspace Diffusion~\cite{diffusionsubspace}
				& Diffusion
				& Subspace-clustering analysis \\
				\midrule
				
				\multirow{3}{*}{\makecell[l]{Clustering and \\ Embedding Learning\\with Flow Matching}}
				& SCFM~\cite{scfm}
				& Flow Matching
				& Structured source coupling \\
				
				& CPFM~\cite{cpfm}
				& Flow Matching
				& Joint embedding and reconstruction \\
				
				& Straight-Path FM~\cite{straightpathfm}
				& Flow Matching
				& Missing-view completion and cluster alignment \\
				\midrule
				
				\multirow{2}{*}{\makecell[l]{Clustering-Guided\\Flow Matching}}
				& Latent-CFM~\cite{latentcfm}
				& Flow Matching
				& Latent-conditioned transport \\
				
				& SubFlow~\cite{subflow}
				& Flow Matching
				& Sub-mode-conditioned transport \\
				\bottomrule\hline
			\end{tabular}
		}
	\end{table*}
	Clustering aims to discover latent groups and semantic structures from unlabeled data. Given samples $\{x_i\}_{i=1}^{N}$, a common pipeline first maps each sample to an embedding $z_i=E(x_i)$ and then groups the embeddings according to their similarities. Clustering performance therefore depends not only on the assignment algorithm, but also on whether the representation space preserves semantic information and removes task-irrelevant variations.
	
	Early deep clustering methods showed the value of jointly learning representations and cluster assignments. Deep Embedded Clustering (DEC)~\cite{dec} progressively refines soft assignments in a learned low-dimensional space, while DeepCluster~\cite{Cluster} alternates between $k$-means clustering and representation learning using cluster assignments as pseudo-labels. These methods established important deep clustering frameworks, but they mainly rely on deterministic embeddings rather than explicit generative processes.
	
	Diffusion and flow-based models provide a different view. Their internal representations change with noise level, time, and network depth. A feature extracted from a generative model can be written as
	\begin{equation}
		x_t=\alpha_t x+\sigma_t\epsilon,
		\qquad
		h_{\ell,t}(x)=H_{\ell}(x_t,t),
	\end{equation}
	
	where $h_{\ell,t}(x)$ denotes the feature extracted from layer $\ell$ at time $t$. Different layers and timesteps may reveal different cluster structures. Cluster information can also be introduced into the latent prior, the assignment process, or the coupling used to train a continuous flow.
	
	Diffusion models and flow-based models have been used to support representation learning, cluster assignment, and latent group discovery. In contrast, applying standard $k$-means to frozen CLIP, DINO, or other foundation-model features is usually treated as a general feature-clustering baseline. Related methods can be organized into four directions: generative clustering with diffusion models, diffusion-based assignment and representation clustering, clustering and embedding learning with Flow Matching, and clustering-guided Flow Matching.
	
	\textbf{Generative Clustering with Diffusion Models} combines cluster discovery and data generation within one probabilistic model. Instead of learning an embedding first and performing clustering afterward, these methods use cluster structure to shape the latent distribution learned by the diffusion model.
	
	ClusterDDPM\cite{ClusterDDPM} combines an expectation--maximization procedure with a conditional denoising diffusion model. In the E-step, it estimates a mixture-of-Gaussians prior over the learned latent representations. In the M-step, the diffusion model learns clustering-friendly representations and matches their distribution to the estimated mixture prior.
	
	This design handles representation learning, cluster assignment, and cluster-conditional generation in one framework. Samples assigned to the same mixture component share a similar latent structure, while different components represent different groups. However, the latent encoder, mixture model, and diffusion generator must be optimized together. The number of clusters is also usually specified before training.
	
	\textbf{Diffusion-Based Assignment and Representation Clustering} uses diffusion to predict cluster assignments or obtain features that are more suitable for clustering. The diffusion process does not always operate in image space. It can instead act on assignment vectors, semantic conditions, or intermediate network features.
	
	Clustering via Self-Supervised Diffusion (CLUDI)~\cite{CLUDI} applies diffusion to the cluster-assignment space. It first extracts image features with a pretrained Vision Transformer. A diffusion-based teacher then produces several stochastic assignment predictions, while a student model combines them into more stable clustering results.
	
	The stochastic reverse process allows CLUDI to represent several possible assignments for an uncertain sample. The teacher--student objective encourages these predictions to remain consistent. This improves clustering stability, although repeated reverse diffusion steps increase the computational cost.
	
	DiFiC~\cite{DiFiC} focuses on fine-grained image clustering. In this setting, small differences between categories can be hidden by changes in background, pose, or texture. Instead of directly clustering high-dimensional image features, DiFiC infers compact textual conditions that explain how each image is generated by a conditional diffusion model.
	
	The inferred conditions are further refined using neighborhood similarity and diffusion-based semantic constraints. The resulting clustering space focuses more on object-level meaning than on raw appearance. This is useful when categories have similar global structures but differ in a small number of semantic details.
	
	Diffusion Embedded Clustering (DiEC)~\cite{diec} directly uses intermediate activations from a pretrained diffusion U-Net. It views features from different layers and timesteps as a representation trajectory. DiEC first selects a suitable network layer and then searches for a timestep that produces clear cluster structure.
	
	After selecting the layer--timestep pair, DiEC applies a lightweight residual mapping to refine the diffusion features. It then uses a DEC-style self-training objective together with graph and entropy regularization. A separate denoising-consistency branch helps preserve the original diffusion representations during clustering.
	
	The relation between diffusion learning and clustering has also been studied theoretically. Under a mixture-of-low-rank-Gaussians assumption and a low-rank denoising model, optimizing the diffusion objective can be related to a subspace-clustering problem~\cite{diffusionsubspace}. This result helps explain why low-dimensional group structure can emerge in diffusion representations even without an explicit clustering objective.
	
	\textbf{Clustering and Embedding Learning with Flow Matching} uses continuous transport to learn structured representations or recover information needed for clustering. Standard Flow Matching learns a velocity field that transports samples from a simple source distribution to the data distribution:
	\begin{equation}
		z_0\sim p_0,
		\qquad
		\frac{\mathrm{d}z_t}{\mathrm{d}t}
		=
		v_{\theta}(z_t,t),
		\qquad
		z_1\sim p_{\mathrm{data}}.
	\end{equation}
	
	When the source is a standard Gaussian distribution, its variables do not usually have a clear connection to semantic groups or cluster identities. Recent methods therefore add structure to the source distribution, learn a low-dimensional embedding together with the flow, or design transport paths that preserve cluster consistency.
	
	Structured Coupling for Flow Matching (SCFM)~\cite{scfm} adds structured latent variables and external transport noise to the source state. The structured variables represent semantic factors or cluster information, while the noise variables provide the remaining randomness needed for generation. A latent-variable model learns the structured prior, and Flow Matching learns the continuous map from this source to the data distribution.
	
	Because the structured variables are introduced before transport begins, cluster information becomes part of the generative process. It is not obtained only by clustering flow features after training. SCFM supports clustering, disentanglement, and downstream prediction while retaining data-generation ability.
	
	Coupled Flow Matching (CPFM)~\cite{cpfm} jointly learns continuous flows in the original data space and a low-dimensional embedding space. The data-to-embedding direction produces compact representations, while the embedding-to-data direction supports reconstruction and generation.
	
	CPFM models the relation between data and embeddings as a conditional distribution rather than using a fixed deterministic encoder. The learned embedding can preserve selected semantic factors and can be used for clustering or visualization. The reverse flow also tests whether the embedding retains enough information to reconstruct the original sample.
	
	Straight-Path Flow Matching for Incomplete Multi-View Clustering~\cite{straightpathfm} applies Flow Matching directly to incomplete multi-view data. In this task, some views of each sample are missing. The method learns a direct transport path from an observed-view representation to the corresponding missing-view representation.
	
	The completed representations are trained together with cluster-level alignment and entropy-based regularization. This encourages different views of the same sample to produce consistent cluster assignments. Unlike diffusion-based completion that starts from unrelated Gaussian noise, the paired flow path directly connects representations from the same sample and is better suited to preserving cluster identity.
	
	\textbf{Clustering-Guided Flow Matching} uses cluster, mode, or latent-group information to simplify the transport problem. In these methods, clustering is not always the final task. Instead, group structure is used to organize transport paths or provide additional conditions to the velocity network.
	
	Efficient Flow Matching using Latent Variables introduces Latent-CFM~\cite{latentcfm}. It uses features extracted by a pretrained latent-variable model to represent the hidden multi-modal structure of the data. These latent features are then provided as conditions to the Flow Matching model.
	
	Latent-CFM avoids learning a single global velocity field without information about the data modes. Samples with similar latent features can follow related transport paths, which simplifies velocity prediction. The latent features can also be used as interpretable conditions for controlled generation. However, the method depends on the quality of the pretrained latent-variable model.
	
	Sub-mode Conditioned Flow Matching (SubFlow)~\cite{subflow} uses clustering to reduce mode loss in one-step generation. A broad class may contain several fine-grained groups, but a flow conditioned only on the class label can average their target velocities and favor the most common modes. SubFlow first performs semantic clustering within each class and assigns a sub-mode index to every sample.
	
	The flow model is then conditioned on the discovered sub-mode rather than only on the broad class label. Each conditional transport problem covers a smaller and more consistent part of the data distribution. This reduces averaging between different modes and improves generation diversity. Although SubFlow is designed mainly for generation, it shows how cluster structure can be used to define more informative conditions for Flow Matching.
	\subsubsection{World Model}\label{World Model}
	
	Ha and Schmidhuber~\cite{ha2018world} first described a world model as an internal model constructed from perception and environmental feedback to predict the future states of the external world. With the recent development of large-scale video and representation models, such as Sora~\cite{brooks2024sora}, V-JEPA 2~\cite{assran2025vjepa2}, Genie~\cite{bruce2024genie}, and Cosmos~\cite{nvidia2025cosmos}, world models are no longer confined to model-based reinforcement learning (MBRL). Instead, they have become increasingly relevant to higher-level intelligent behaviors, including state estimation, future prediction, planning, decision making, and interactive control. In embodied intelligence, such models can serve as learned world simulators, substantially reducing the cost of interacting with the physical world and providing a core substrate for broader progress toward general-purpose intelligence.
	
	From the perspective of representation learning, a world model can be viewed as an application of representation learning to dynamic environment modeling. It compresses high-dimensional and complex observations into latent states with semantic information and predictive capability, and then learns the evolution dynamics of the environment in this latent space. Given observations, actions, trajectories, or other conditioning signals, world models aim to predict future visual states, motion states, or more abstract representations of world evolution.
	
	In this review, we further decompose the broad concept of world models in relation to diffusion- and flow-matching-based representation learning. We focus on three major directions: video world models, 3D/4D world models, and robotics and embodied world models. These directions differ in the representation spaces they use, but they share the common goal of modeling future states from existing perceptual information through image, video, 3D, or embodied representations.
	
	\paragraph{Video World Models}
	
	Video can be regarded as a projection of world states into pixel space, and the abundance of large-scale visual data, together with mature video generation techniques, makes video generation an important route toward world modeling.
	One direct approach is to construct video world models either in pixel space or in VAE latent spaces designed for reconstruction and generation. Such methods typically learn
	\begin{equation}
		p(o_{t+1:t+H} \mid o_{\leq t}, c),
	\end{equation}
	where the condition \(c\) may denote actions, camera trajectories, language instructions, or navigation goals. GameNGen demonstrates that diffusion models can act as real-time neural game engines by autoregressively generating next frames in DOOM from past frames and actions, while using conditioning augmentation to mitigate long-horizon drift~\cite{valevski2024gamengen}. DIAMOND further shows that, on Atari 100k, a pixel-space diffusion world model that preserves visual details can improve model-based RL agent training and can be extended into an interactive neural game engine~\cite{alonso2024diamond}. At a larger scale, the Cosmos World Foundation Model platform combines large-scale video curation, video tokenization, diffusion/autoregressive world models, and Physical AI adaptation, making video generation models fine-tunable world priors for robotics and autonomous driving~\cite{nvidia2025cosmos}. These works suggest that the strong visual fidelity of diffusion-based video generation and internet-scale video priors make them well suited for modeling pixel-level projections of world states. However, their limitations are also clear: high inference cost, long-horizon error accumulation, and the fact that scaling alone remains insufficient for ensuring physical adherence.
	
	Another line of work performs world modeling in latent representation space. Unlike pixel-level or token-level reconstruction, these methods predict future states in representation space, encouraging the model to learn semantically abstract features and internal dynamics of the environment. Representative approaches include the family of joint-embedding predictive architectures (JEPAs), such as I-JEPA~\cite{assran2023ijepa}, MC-JEPA~\cite{bardes2023mcjepa}, V-JEPA~\cite{bardes2024vjepa}, AD-L-JEPA~\cite{zhu2025adljepa}, and V-JEPA 2~\cite{assran2025vjepa2}. These methods are computationally efficient; for example, V-JEPA 2 demonstrates more efficient inference and planning than generative video world models~\cite{assran2025vjepa2}. Meanwhile, JEDI introduces a diffusion objective into the JEPA framework, learning future embeddings through denoising in latent space, thereby avoiding the high cost of pixel-space diffusion and achieving faster sampling with lower VRAM consumption on Atari 100k~\cite{lim2026jedi}. In addition, some work applies flow matching directly in the representation space of vision foundation models rich in depth information. For instance, GLD repurposes the feature space of geometric foundation models for novel view synthesis and combines it with RAE-based RGB reconstruction to improve 3D consistency and training efficiency~\cite{jang2026gld}.
	
	A key question in this direction is that not all latent spaces are suitable for diffusion-based generation. The analysis of diffusion-friendly latent manifolds in PAE indicates that a useful latent space requires not only reconstruction fidelity but also coherent spatial structure, local manifold continuity, and global semantics~\cite{yue2026pae}. Therefore, in world modeling, the quality and distribution of learned representations directly determine whether future prediction can be both compact and generative.
	
	\paragraph{3D/4D World Models}
	
	For autonomous driving and embodied robotics, generating RGB video alone is often insufficient for geometry-aware reasoning. As a result, 3D/4D representations such as RGB-D, occupancy, 3D Gaussian splatting (3DGS), and point clouds have become important representation forms. Compared with 2D video, 3D/4D world models place greater emphasis on geometric and temporal consistency, enabling more accurate and verifiable future-state prediction. Representative works include UniScene~\cite{li2024uniscene}, DrivingSphere~\cite{yan2024drivingsphere}, X-Scene~\cite{yang2025xscene}, COME~\cite{shi2025come}, R2Flow~\cite{nakashima2024r2flow}, LidarDM~\cite{zyrianov2024lidardm}, HoloDrive~\cite{wu2024holodrive}, and LOGen~\cite{kirby2024logen}.
	
	OccSora uses a 4D scene tokenizer and a diffusion transformer to generate trajectory-conditioned occupancy videos, pushing world simulation from 2D video toward 4D semantic occupancy~\cite{wang2024occsora}. HY-World 2.0 constructs a complete pipeline consisting of panorama generation, trajectory planning, view generation, feed-forward reconstruction, and 3DGS rendering to generate interactive and navigable 3D worlds~\cite{hyworld2026}. These methods have the advantage of verifiable geometry and can interface with rendering or physics engines, making them better suited to safety-critical scenarios such as autonomous driving and robotics. At the same time, they impose higher requirements on datasets, annotation, training cost, and inference cost.
	
	\paragraph{Robotics and Embodied World Models}
	
	In robotics and embodied AI, the role of world models is expanding from ``future video generation'' to policy learning, policy evaluation, and representation priors for vision-language-action (VLA) models. IRASim generates real-robot videos conditioned on action trajectories, providing a diffusion-based implementation of a robot action simulator~\cite{zhu2024irasim}. WorldEval and dWorldEval use world models as scalable policy evaluators: the former predicts robot execution outcomes through latent actions, while the latter performs prediction in a discrete diffusion token space, both aiming to reduce reliance on expensive real-world rollouts~\cite{li2025worldeval,li2026dworldeval}. More recent work in 2026 emphasizes tighter coupling with VLA models. World2Act post-trains VLA models with world-model video-dynamics latents, reducing dependence on pixel-level rollout artifacts~\cite{vuong2026world2act}. LaMP injects 3D scene flow generated by a flow-matching Motion Expert as a latent motion prior into an Action Expert, improving VLA robustness under unfamiliar spatial dynamics~\cite{wang2026lamp}. LaWAM replaces full future video generation with latent visual subgoals, reducing latency while retaining dynamics-aware control~\cite{chen2026lawam}. These trends suggest that, in generalist embodied applications, a world model is not merely a visual simulator but an intermediate layer connecting representation learning, future prediction, and embodied policy.
	
	Overall, diffusion and flow-based world models are pushing representation learning from discriminative perception toward predictive and controllable representations. Pixel/video world models are strong in visualization, data flywheels, and general-purpose video priors; latent and JEPA-style world models emphasize efficiency, planning, and task abstraction; and 3D/4D world models strengthen geometric consistency and physical grounding. A central challenge for future work is to unify high-fidelity generation, long-horizon consistency, action controllability, physical plausibility, and efficient inference, so that world models can become reusable world interfaces for generalist embodied agents.
	
	\subsubsection{Unified Model}\label{Unified Model}
	\begin{table*}[t!]
		\centering
		\tiny
		\setlength{\tabcolsep}{1.8pt}
		\renewcommand{\arraystretch}{1.12}
		\caption{Representative unified multimodal models with diffusion or flow-based generation components.}
		\label{tab:unified_models}
		\begin{tabularx}{\textwidth}{@{}>{\raggedright\arraybackslash}p{0.15\textwidth}>{\centering\arraybackslash}p{0.14\textwidth}>{\centering\arraybackslash}p{0.15\textwidth}>{\centering\arraybackslash}p{0.08\textwidth}>{\centering\arraybackslash}p{0.05\textwidth}>{\centering\arraybackslash}p{0.12\textwidth}>{\centering\arraybackslash}p{0.09\textwidth}>{\centering\arraybackslash}p{0.14\textwidth}@{}}
			\toprule
			\textbf{Model} & \textbf{Paradigm} & \textbf{Backbone} & \textbf{Text Enc.} & \textbf{Text Dec.} & \multicolumn{2}{c}{\textbf{Visual Enc.}} & \textbf{Visual Dec.}\\
			\cmidrule(lr){6-7}
			& & & & & \textbf{Und.} & \textbf{Gen.} &\\
			\midrule
			\multicolumn{8}{@{}l}{\textit{Generalist LLM/VLMs with diffusion or flow decoders }} \\
			\midrule
			DreamLLM~\cite{dong2023dreamllm} & Cont. AR, Diff & LLaMA & \multicolumn{2}{c}{LLaMA} & \multicolumn{2}{c}{OpenAI-CLIP} & SD-2.1\\
			SEED~\cite{ge2023planting} & Cont. AR, Diff & OPT & \multicolumn{2}{c}{OPT} & \multicolumn{2}{c}{SEED-Tok.(Query)} & SD\\
			SEED-LLaMA~\cite{ge2024making} & Cont. AR, Diff & LLaMA-2/Vicuna & \multicolumn{2}{c}{LLaMA-2/Vicuna} & \multicolumn{2}{c}{SEED-Tok.(Query)} & unCLIP-SD\\
			SEED-X~\cite{ge2024seed} & Cont. AR, Diff & LLaMA-2 & \multicolumn{2}{c}{LLaMA-2} & \multicolumn{2}{c}{SEED-Tok.(Query)} & SDXL\\
			Emu2~\cite{sun2024generative} & Cont. AR, Diff & LLaMA & \multicolumn{2}{c}{LLaMA} & \multicolumn{2}{c}{EVA-CLIP} & SDXL\\
			MM-Interleaved~\cite{tian2024mm} & Cont. AR, Diff & Vicuna & \multicolumn{2}{c}{Vicuna} & \multicolumn{2}{c}{OpenAI-CLIP} & SD-v2.1\\
			PUMA~\cite{fang2024puma} & Cont. AR, Diff & LLaMA-3 & \multicolumn{2}{c}{LLaMA-3} & \multicolumn{2}{c}{OpenAI-CLIP} & SDXL\\
			Unifluid~\cite{fan2025unified} & Cont. AR, Diff & Gemma-2 & \multicolumn{2}{c}{Gemma-2} & SigLIP & SD-VAE & Diffusion MLP\\
			BLIP3-o~\cite{chen2025blip3} & Cont. AR, Diff & Qwen2.5-VL & \multicolumn{2}{c}{Qwen2.5-VL} & OpenAI-CLIP & Query & Lumina-Next \\
			OmniGen2~\cite{wu2025omnigen2} & Cont. AR, Diff & Qwen2.5-VL & \multicolumn{2}{c}{Qwen2.5-VL} & \multicolumn{2}{c}{SigLIP} & OmniGen \\
			Ovis-U1~\cite{wang2025ovis} & Cont. AR, Diff & Ovis & \multicolumn{2}{c}{Ovis} & \multicolumn{2}{c}{AimV2} & MMDiT \\
			UniCode$^2$~\cite{chan2025unicode2} & Cont. AR, Diff/Flow & Qwen2.5 & \multicolumn{2}{c}{Qwen2.5} & \multicolumn{2}{c}{SigLIP+RQ} & FLUX.1-dev / SD-1.5 \\
			Nexus-Gen~\cite{zhang2025nexus} & Cont. AR, Flow & Qwen2.5-VL & \multicolumn{2}{c}{Qwen2.5-VL} & Qwen2.5-VL ViT & Query & FLUX \\
			X-Omni~\cite{geng2025x} & Cont. AR, Flow & Qwen2.5-VL & \multicolumn{2}{c}{Qwen2.5-VL} & Qwen2.5-VL ViT & SigLIP & FLUX \\
			Qwen-Image~\cite{wu2025qwen} & Cont. AR, Diff & Qwen2.5-VL & \multicolumn{2}{c}{Qwen2.5-VL} & \multicolumn{2}{c}{Qwen2.5-VL ViT} & MMDiT \\
			UniPic-2.0~\cite{wei2025skywork} & Cont. AR, Diff & Qwen2.5-VL & \multicolumn{2}{c}{Qwen2.5-VL} & Qwen2.5-VL ViT & Query & SD3.5 \\
			MammothModa2~\cite{shen2025mammothmoda2} & Cont. AR, Diff & Qwen3-VL & \multicolumn{2}{c}{Qwen3-VL} & Qwen3-VL ViT & MammothTok & Single-stream DiT \\
			UniAR~\cite{pengunified} & Cont. AR, Diff &  Qwen3 & \multicolumn{2}{c}{ Qwen3} & \multicolumn{2}{c}{SigLiP2} & SD3.5-DiT \\
			HyperCLOVA X~\cite{team2026hyperclova} & Cont. AR, Diff & HyperCLOVA X & \multicolumn{2}{c}{HyperCLOVA X} & Qwen2.5-VL ViT & TA-Tok & FLUX-MMDiT \\
			\midrule
			\multicolumn{8}{@{}l}{\textit{End-to-end diffusion or flow-style}} \\
			\midrule
			VersatileDiffusion~\cite{xu2023versatile} & Cont. Diff. & UNet & CLIP-Text & GPT2 & CLIP-Image & SD-VAE & SD-VAE \\
			UniDiffuser~\cite{bao2023one} & Cont. Diff. & U-ViT & CLIP-Text & GPT-2 & CLIP-Image & SD-VAE & SD-VAE \\
			Dual Diffusion~\cite{li2024dual} & Cont. Diff. & D-DiT & \multicolumn{2}{c}{D-DiT} & \multicolumn{2}{c}{SD-VAE} & SD-VAE \\
			UniModel~\cite{zhang2025unimodel} & Cont. Diff. & Qwen-Image & \multicolumn{2}{c}{---} & \multicolumn{2}{c}{Qwen-Image VAE} & Qwen-Image VAE \\
			FUDOKI~\cite{wang2025fudoki} & Dis. Flow & DeepSeek-LLM & \multicolumn{2}{c}{DeepSeek-LLM} & SigLIP & VQGAN & VQGAN \\
			NExT-OMNI~\cite{luo2025next} & Dis. Flow & Qwen2.5 & \multicolumn{2}{c}{Qwen2.5} & \multicolumn{2}{c}{VQVAE } & VQVAE \\
			UniDisc~\cite{swerdlow2025unified} & Dis. Diff. & DiT & \multicolumn{2}{c}{DiT} & \multicolumn{2}{c}{MAGVIT-v2} & MAGVIT-v2 \\
			Lavida-O~\cite{li2025lavidao} & Dis. Diff. & LaViDa & \multicolumn{2}{c}{LaViDa} & SigLIP & VQ-Encoder & VQ-Encoder \\
			MMaDA~\cite{yang2025mmada} & Dis. Diff. & LLaDA & \multicolumn{2}{c}{LLaDA} & \multicolumn{2}{c}{MAGVIT-v2} & MAGVIT-v2 \\
			Lumina-DiMOO~\cite{xin2025lumina} & Dis. Diff. & LLaDA & \multicolumn{2}{c}{LLaDA} & \multicolumn{2}{c}{aMUSEd-VQ} & aMUSEd-VQ \\
			Muddit~\cite{shi2025muddit} & Dis. Diff. & MM-DiT & \multicolumn{2}{c}{MM-DiT} & \multicolumn{2}{c}{VQGAN} & VQGAN \\
			LLaDA2.0-Uni~\cite{ai2026llada2} & Dis. Diff. & LLaDA2.0 & \multicolumn{2}{c}{LLaDA2.0} & \multicolumn{2}{c}{SigLIP-VQ} & SigLIP-VQ \\
			Omni-Diffusion~\cite{li2026omni} & Dis. Diff. & Dream & \multicolumn{2}{c}{Dream} & \multicolumn{2}{c}{MAGVIT-v2} & MAGVIT-v2 \\
			Dynin-Omni~\cite{kim2026dynin} & Dis. Diff. & MMaDA & \multicolumn{2}{c}{MMaDA} & \multicolumn{2}{c}{MAGVIT-v2} & MAGVIT-v2 \\
			ViewMask-1-to-3~\cite{zhu2026viewmask} & Dis. Diff. & LLaDA & \multicolumn{2}{c}{LLaDA} & \multicolumn{2}{c}{MAGVIT-v2} & MAGVIT-v2 \\
			\midrule
			\multicolumn{8}{@{}l}{\textit{Fused autoregressive--diffusion or autoregressive--flow}} \\
			\midrule
			Transfusion~\cite{zhou2025transfusion} & Cont. AR+Diff & LLaMA-2 & \multicolumn{2}{c}{LLaMA-2} & \multicolumn{2}{c}{SD-VAE} & SD-VAE \\
			MonoFormer~\cite{zhao2024monoformer} & Cont. AR+Diff & TinyLLaMA & \multicolumn{2}{c}{TinyLLaMA} & \multicolumn{2}{c}{SD-VAE} & SD-VAE \\
			LMFusion~\cite{shi2024llamafusion} & Cont. AR+Diff & LLaMA & \multicolumn{2}{c}{LLaMA} & \multicolumn{2}{c}{SD-VAE+UNet down.} & SD-VAE+UNet up. \\
			TUNA~\cite{liu2025tuna} & Cont. AR+Diff & Qwen-2.5 & \multicolumn{2}{c}{Qwen-2.5} & SigLIP2 & VAE & VAE \\
			JanusFlow~\cite{ma2024janusflow} & Cont. AR+Flow & DeepSeek-LLM & \multicolumn{2}{c}{DeepSeek-LLM} & SigLIP & SDXL-VAE & SDXL-VAE \\
			BAGEL~\cite{deng2025bagel} & Cont. AR+Flow & Qwen2.5 & \multicolumn{2}{c}{Qwen2.5} & SigLIP2 & FLUX-VAE & FLUX-VAE \\
			Mogao~\cite{liao2025mogao} & Cont. AR+Flow & Qwen2.5 & \multicolumn{2}{c}{Qwen2.5} & SigLIP & SDXL-VAE & SDXL-VAE \\
			LightFusion~\cite{wang2025lightfusion} & Cont. AR+Flow & Qwen2.5-VL+Wan2.2 & \multicolumn{2}{c}{Qwen2.5-VL} & Qwen2.5-VL ViT & Wan2.2-TI2V & DCAE \\
			HBridge~\cite{wang2025hbridge} & Cont. AR+Flow & Qwen2.5+OmniGen2 & \multicolumn{2}{c}{Qwen2.5-VL} & Qwen2.5-VL ViT & SigLIP & OmniGen \\
			EMMA~\cite{he2025emma} & Cont. AR+Flow & Qwen3 & \multicolumn{2}{c}{Qwen3} & SigLIP & DCAE & DCAE \\
			Show-o~\cite{xie2024show} & Dis. AR+Diff & LLaVA-v1.5-Phi & \multicolumn{2}{c}{Phi} & \multicolumn{2}{c}{MAGVIT-v2} & MAGVIT-v2 \\
			UniCTokens~\cite{an2026unictokens} & Dis. AR+Diff & Show-o & \multicolumn{2}{c}{Phi} & \multicolumn{2}{c}{MAGVIT-v2} & MAGVIT-v2 \\
			\bottomrule
		\end{tabularx}
	\end{table*}
	The core objective of a unified multimodal model is to perform visual understanding and generation within a single architecture and shared parameters, taking multimodal inputs (text, images, video) and producing outputs in one or more modalities. Such models generally decompose into three components: a \textit{modality-specific encoder} mapping heterogeneous inputs into a unified representation space (e.g., CLIP~\cite{radford2021learning}, SigLIP~\cite{zhai2023sigmoid}, or discrete tokenizers such as VQGAN~\cite{esser2021taming}); a \textit{fusion backbone} for cross-modal interaction and reasoning, typically an autoregressive LLM such as LLaMA~\cite{touvron2023llama} or Qwen(-VL)~\cite{Bai2025Qwen25VLTR}, or a diffusion language model such as LLaDA~\cite{nie2025llada}; and a \textit{modality-specific decoder} mapping fused representations back to the target modality, commonly a diffusion or flow-based image decoder such as Stable Diffusion~\cite{StableDiffusion} or FLUX~\cite{blackforestlabs2024flux}.
	
	Where the diffusion or flow mechanism is introduced---encoder, backbone, or decoder---is the key factor distinguishing unified model architectures. As this survey focuses on diffusion/flow-related representation learning, we restrict our discussion to unified models whose generation pathway includes a diffusion or flow component. Purely autoregressive models, such as Emu3~\cite{wang2024emu3} and Janus(-Pro)~\cite{wu2025janus, chen2025januspro}, instead use discrete-token decoders (e.g., VQGAN~\cite{esser2021taming}) and next-token prediction without any denoising process, and are not discussed further. As summarized in Table~\ref{tab:unified_models}, existing work falls into three paradigms based on how the diffusion/flow module couples with the backbone: (i) \textit{cascaded} models, i.e., LLMs/VLMs equipped with diffusion or flow decoders; (ii) \textit{end-to-end diffusion- or flow-style} models; and (iii) \textit{fused} autoregressive--diffusion/flow models.
	
	\paragraph{LLM/VLM-Driven Diffusion/Flow Decoders.}
	
	This category of methods builds upon mature LLM/VLM technologies, combining an existing understanding backbone with an external generative head. A language or multimodal backbone, such as LLaMA or Qwen2.5-VL, handles semantic planning and token-level autoregressive modeling, while pixel-level synthesis is delegated to an independent diffusion or flow decoder, with the two connected via conditioning embeddings or continuous tokens.
	
	Early methods, including DreamLLM~\cite{dong2023dreamllm} and the SEED series~\cite{ge2023planting,ge2024making,ge2024seed}, use CLIP-family encoders for image semantics and diffusion decoders (e.g., Stable Diffusion) for pixel reconstruction, often sharing a single tokenizer between understanding and generation branches. Later methods such as Emu2~\cite{sun2024generative} and PUMA~\cite{fang2024puma} pair CLIP-family encoders with SDXL decoders to extend interleaved image--text generation.
	
	As specialized visual foundation models matured, recent methods decouple the encoders used for understanding and generation. Unifluid~\cite{fan2025unified}, for instance, employs SigLIP for understanding and SD-VAE for generative encoding, while lightweight generative heads (e.g., diffusion MLPs) reduce interference between the two representations. BLIP3-o~\cite{chen2025blip3} and OmniGen2~\cite{wu2025omnigen2} adopt Qwen2.5-VL as a shared backbone with stronger diffusion/flow decoders such as OmniGen~\cite{xiao2025omnigen} and MMDiT~\cite{esser2024scaling}.
	
	More recently, methods such as Nexus-Gen~\cite{zhang2025nexus} and X-Omni~\cite{geng2025x} replace conventional diffusion objectives with flow matching paired with FLUX-family decoders, reflecting the growing efficiency and quality advantages of flow-based modeling over DDPM-style diffusion. The latest methods, including Qwen-Image~\cite{wu2025qwen} and UniPic-2.0~\cite{wei2025skywork}, increasingly converge on Qwen-ViT as the understanding-side encoder, while decoder architectures gradually shift from UNet-based diffusion to single-stream DiT designs.
	
	\paragraph{End-to-End Diffusion/Flow Models}
	Unlike cascaded paradigms, the end-to-end paradigm dispenses with an independent autoregressive language backbone, instead jointly modeling textual and visual representations within a unified diffusion or flow process. According to the modeling space, these methods fall into two major routes: \textbf{continuous diffusion} and \textbf{discrete diffusion}.
	
	Early continuous-domain methods, such as VersatileDiffusion~\cite{xu2023versatile} and UniDiffuser~\cite{bao2023one}, encode text and images separately via CLIP-Text and CLIP-Image but share a UNet or U-ViT backbone for joint cross-modal diffusion. Later work, such as UniModel~\cite{zhang2025unimodel}, further unifies textual and visual representations within the same DiT architecture and VAE latent space, weakening modality-specific encoder designs and advancing multimodal modeling from sharing a denoising network toward sharing both the representation space and the generative process.
	
	Discrete diffusion has also developed rapidly, led by the Masked Diffusion Model (MDM), which has emerged as an important architecture for discrete generation owing to its bidirectional modeling capability, architectural simplicity, high parallelism, and favorable scaling properties. This line of research first produced diffusion large language models (dLLMs), such as LLaDA~\cite{nie2025llada} and Dream~\cite{ye2025dream}, and was subsequently extended to multimodal settings through models such as LLaDA-V~\cite{you2026llada} and Dimple~\cite{yu2025dimple}. A central idea in discrete-domain Unified Models is to discretize visual signals into visual tokens using vector-quantized tokenizers such as VQGAN, enabling images and text to be modeled within a unified token space; image generation can then reuse mask-based discrete diffusion and share the same masking, denoising, and token-prediction mechanisms with text generation, as in MMaDA~\cite{yang2025mmada} and Lumina-DiMOO~\cite{xin2025lumina}. ViewMask-1-to-3~\cite{zhu2026viewmask} further extends this formulation to multi-view generation, demonstrating the applicability of the MDM architecture to more structurally constrained generation tasks.
	
	Beyond mask-based discrete diffusion, recent studies have also explored discrete flow matching (DFM). FUDOKI~\cite{wang2025fudoki} models generative trajectories in discrete spaces via metric-induced probability paths, while NExT-OMNI extends DFM to any-to-any omnimodal modeling, understanding, and retrieval across text, image, video, and audio.
	
	\paragraph{Hybrid Autoregressive and Diffusion/Flow Modeling}
	The hybrid paradigm aims to accommodate both autoregressive next-token prediction and diffusion or flow denoising objectives within a single Transformer, typically by applying different attention masks or learning objectives to tokens from different modalities, so that text is generated autoregressively while images are generated via diffusion or flow.
	
	Transfusion~\cite{zhou2025transfusion} and MonoFormer~\cite{zhao2024monoformer} are foundational works in this direction, both incorporating continuous-domain diffusion objectives directly into LLaMA-based backbones with SD-VAE for visual encoding and decoding. In the flow-matching direction, JanusFlow~\cite{ma2024janusflow} is among the first to introduce a flow-based generation objective into the hybrid framework. BAGEL~\cite{deng2025bagel} and Mogao~\cite{liao2025mogao} further adopt Qwen2.5 as the backbone, SigLIP2 as the understanding encoder, and FLUX-VAE or SDXL-VAE as the generative encoder/decoder, forming a relatively mature architectural combination.
	
	Recent methods further improve efficiency and representation decoupling. LightFusion~\cite{wang2025lightfusion} introduces the video-generation foundation model Wan2.2-TI2V as the generative encoder and DCAE as the decoder, exploring unified image and video representations. In addition, Show-o~\cite{xie2024show} represents a discrete-domain extension of the hybrid paradigm (denoted Dis. AR+Diff.), using a MAGVIT-v2 tokenizer to unify autoregressive text generation with discrete diffusion-based image generation under a shared mask-based training objective, further blurring the boundary between autoregressive and diffusion modeling.
	
	Overall, the three paradigms exhibit several clear evolutionary trends from the perspective of representation learning: (1) Diffusion- and flow-based denoising objectives are developing in parallel and increasingly competing across different model families. (2) Because understanding and generation require representations at different levels of granularity~\cite{wu2025janus}, the choice between unified and decoupled representation spaces has become a central architectural distinction in Unified Models, reflecting the trade-off between semantic abstraction and visual detail preservation.
	
	\section{Discussion and Future Directions}
	Despite the significant potential that diffusion models and flow-based models have demonstrated in representation learning, this field remains in its early stages of rapid development, with many open problems warranting further exploration.
	
	First, current research heavily favors diffusion models over flow-based models. Flow-based models, with their deterministic probability paths and more efficient inference, may offer unique advantages in the efficiency and controllability of representation learning, yet relevant explorations remain scarce. We believe that systematically incorporating the distinctive characteristics of flow-based models, such as temporal asymmetry and vector field geometry, into representation learning is a direction worth pursuing.
	
	Second, most existing work evaluates representation quality through indirect downstream task metrics, lacking a unified benchmark for the intrinsic properties of representations themselves, such as linear separability, disentanglement, and interpretability. Establishing a standardized evaluation framework for generative model representations would facilitate fairer comparisons across methods and guide model design.
	
	Third, the scope of generative model representations is gradually expanding from perception tasks toward higher-level intelligent tasks including decision-making, planning, and interaction. How to equip representations not only with semantic abstraction capabilities but also with the capacity to encode causal structures, physical laws, and temporal dynamics remains a critical challenge for achieving general-purpose embodied intelligence.
	
	Finally, with the widespread adoption of diffusion models and flow-based models in unified multimodal frameworks, representation learning will play an increasingly central role in the deep integration of understanding and generation. How to reconcile the differing demands of these two task families on representations within a shared parameter architecture, including the tension between semantic abstraction and detail fidelity, will be a core issue for unified models.
	
	We look forward to continued progress in this field that can truly bridge the gap between generation and understanding, making generative models a foundational component of general-purpose intelligent systems.
	
	\section{Declarations}
	\subsection{Funding}
	Not applicable.
	\subsection{Conflicts of Interest}
	The authors declare that they have no competing financial or non-financial interests.
	\subsection{Data Availability}
	Not applicable.
	\subsection{Code Availability}
	Not applicable.
	\subsection{Authors' Contributions}
	The completion of this paper was a result of the collaborative efforts of all authors. All authors contributed to the writing and editing of this review. Hongyuan Zhang also handled project administration.
	\subsection{Acknowledgements}
	Not applicable.

	
	\bibliography{citations}
	
\end{document}